\pdfoutput=1
\documentclass[11pt]{article}
\usepackage{authblk}
\usepackage[]{acl}

\usepackage{times}
\usepackage{latexsym}

\usepackage[T2A,T1]{fontenc}
\usepackage[utf8]{inputenc}
\usepackage[greek,russian,english]{babel}

\usepackage{microtype}
\usepackage{inconsolata}
\usepackage{graphicx}
\usepackage{microtype}
\usepackage{amsmath}
\usepackage{graphicx}
\usepackage{subfigure}
\usepackage{amsfonts}
\usepackage{multirow}
\usepackage{listings}
\usepackage{url}
\usepackage{CJKutf8}
\usepackage{graphicx}
\usepackage{subfigure}
\usepackage{supertabular}
\usepackage{subfigure}
\usepackage{dcolumn,booktabs}
\usepackage{tikz}
\usepackage{xspace}

\def\dnrm#1{\mbox{$_{\hbox{\scriptsize #1}}$}}

\title{\textsc{TransMI}: A Framework to Create Strong Baselines from Multilingual Pretrained Language Models for Transliterated Data}

 \author{{\bf Yihong Liu}, {\bf Chunlan Ma}, {\bf Haotian Ye}, and {\bf Hinrich Sch\"utze}\\
\\
Center for Information and Language Processing, LMU Munich \\ 
 Munich Center for Machine Learning (MCML)\\
\texttt{\{yihong, chunlan, yehao\}@cis.lmu.de}}

\def\secref#1{\S\ref{sec:#1}}
\def\seclabel#1{\label{sec:#1}}

\newcounter{notecounter}
\newcommand{\enotesoff}{\long\gdef\enote##1##2{}}
\newcommand{\enoteson}{\long\gdef\enote##1##2{{
\stepcounter{notecounter}
{\large\bf
\hspace{1cm}\arabic{notecounter} $<<<$ ##1: ##2
$>>>$\hspace{1cm}}}}}
\enoteson
\enotesoff

\begin{document}

\def\frameworkname{\textsc{TransMI}\xspace}
\def\modelname{\textsc{TransMI}\xspace}
\maketitle
\begin{abstract}

% Transliterating related languages that use different scripts into a common script is effective for improving crosslingual transfer in downstream tasks. However, this methodology often makes pretraining a model from scratch unavoidable, as transliteration brings about new subwords not covered in existing multilingual pretrained language models (mPLMs).  This is undesirable because it requires a large computation budget. A more promising way is to make full use of available mPLMs.  To this end, this paper proposes a simple but effective framework: $\textbf{Trans}$literate-$\textbf{M}$erge-$\textbf{I}$nitialize ($\textbf{TransMI}$). TransMI is a strong baseline well-suited for data that is transliterated into a common script by exploiting an mPLM and its tokenizer. TransMI has three stages: ($\textbf{a}$) transliterate the vocabulary of an mPLM into a common script; ($\textbf{b}$) merge the new vocabulary with the original vocabulary; and ($\textbf{c}$) initialize the embeddings of the new subwords.  We apply TransMI to three strong recent mPLMs. Our experiments demonstrate that TransMI not only preserves the mPLM's ability to handle non-transliterated data, but also enables it to effectively process transliterated data, thereby facilitating crosslingual transfer. The results show consistent improvements of 3\% to 34\% for different mPLMs and tasks.  We will make our code and models publicly available.

Transliterating related languages that use different scripts
into a common script is effective for improving
crosslingual transfer in downstream tasks. However, this
methodology often makes pretraining a model from scratch
unavoidable, as transliteration brings about new subwords
not covered in existing multilingual pretrained language
models (mPLMs).  This is undesirable because it requires a large
computation budget. A more promising way
is to make full use of available mPLMs.  To this end, this
paper proposes a simple but effective
framework: \textbf{Trans}literate-\textbf{M}erge-\textbf{I}nitialize
(\textbf{\textsc{TransMI}}). \textsc{TransMI} can create strong baselines for data that is transliterated into a
common script by exploiting an existing mPLM and its
tokenizer without any training.\footnote{Throughout this paper we will simply use
mPLM to refer to both the model and its tokenizer for
convenience.} \textsc{TransMI} has three stages:
(\textbf{a}) transliterate the vocabulary of an mPLM into a
common script; (\textbf{b}) merge the new vocabulary with
the original vocabulary; and (\textbf{c}) initialize the
embeddings of the new subwords.  We apply \textsc{TransMI}
to three strong recent mPLMs. Our experiments
demonstrate that \textsc{TransMI} not only preserves the mPLM's
ability to handle non-transliterated data, but also enables
it to effectively process transliterated data,
thereby facilitating crosslingual transfer across scripts.
The results show consistent improvements of 3\% to 34\% for
different mPLMs and tasks.  We make our code
and models publicly available at \url{https://github.com/cisnlp/TransMI}.

\end{abstract}

\section{Introduction}

Crosslingual transfer refers to applying knowledge gained from one language to the learning or processing of another language \citep{zoph-etal-2016-transfer,wu-dredze-2019-beto,artetxe-etal-2020-cross}. This transfer is attractive as we often do not have enough training data for low-resource languages while training data for high-resource languages is generally abundant \citep{magueresse2020low,hedderich-etal-2021-survey,liu2022effective}. Although recent mPLMs have made remarkable progress in improving crosslingual transfer, they often cannot achieve strong performance when transferring to a wide spectrum of low-resource languages. Lexical overlap, i.e., the phenomenon where vocabularies are shared between languages, is a key factor influencing the quality of crosslingual transfer \citep{pires-etal-2019-multilingual,lin-etal-2019-choosing}. However, because languages are written in different writing systems, or \emph{scripts}, lexical overlap cannot be fully exploited.
% For example, ``\foreignlanguage{russian}{бар}'' (``bar'' in Ukrainian) is exactly the same as ``bar'' in English both in their meaning and after a char-by-char transliteration, but they are regarded as different words that have \emph{no} lexical overlap at all.

\begin{table}
\setlength{\belowcaptionskip}{-0.4cm}
\footnotesize
\centering
\setlength{\tabcolsep}{0.8mm}{}
\begin{tabular}{lr}
\toprule
% \toprule
{\bf original sentence}: & \begin{CJK}{UTF8}{gbsn}今天是个好天气\end{CJK} \\
{\bf transliteration}: & jintianshigehaotianqi\\
\midrule
\multirow{2}{*}{\bf original tokenizer} & \begin{CJK}{UTF8}{gbsn}`\_今天', `是个', `好', `天气'
\end{CJK} \\
  & `\_jint', `ian', `shig', `ehao', `tian', `qi' \\
\midrule
\multirow{2}{*}{\bf modified tokenizer} & \begin{CJK}{UTF8}{gbsn}`\_今天', `是个', `好', `天气'
\end{CJK} \\
 & `\_jintian', `shige', `hao', `tianqi'\\
\bottomrule
% \bottomrule
\end{tabular}
\caption{Tokenization results of a sentence written in its original script (Hani) and its Latin transliteration.\protect\footnotemark~The correct word correspondences are:
(\begin{CJK}{UTF8}{gbsn}今天\end{CJK} – jintian – <today>),
(\begin{CJK}{UTF8}{gbsn}是个\end{CJK} – shige – <is>),
(\begin{CJK}{UTF8}{gbsn}好\end{CJK} – hao – <good>),
(\begin{CJK}{UTF8}{gbsn}天气\end{CJK} – tianqi –
<weather>). The original tokenizer produces nonsensical
strings that do not correspond to the meaning-bearing units. The modified tokenizer correctly tokenizes the transliterated text while also preserving the ability to handle the original sentence. 
% in its original Hani script.
}\label{tab:example_tokenization}
\end{table}

% from the original tokenizer and the tokenizer modified by \frameworkname.

\footnotetext{Hani script generally stands for the Chinese characters. This script can be further classified into Hant (traditional Chinese characters) and Hans (simplified Chinese characters).}

To tackle this problem, a few recent works attempt to apply rule-based transliteration tools and convert all data to a common script \citep{dhamecha-etal-2021-role,muller-etal-2021-unseen,moosa-etal-2023-transliteration}. By doing this, the script diversity no longer poses difficulty in improving lexical overlap, therefore better performance can be obtained. However, this approach either
requires training a brand-new mPLM from
scratch \citep{dhamecha-etal-2021-role,moosa-etal-2023-transliteration}
or involves (parameter-efficient) parameter updates to adapt
the transliterated
data \citep{muller-etal-2021-unseen,purkayastha-etal-2023-romanization},
as the embeddings of the new subwords generated from
transliteration need to be properly trained before a model
can be applied to any downstream tasks. This inevitably
demands a high computing budget. Additionally, such
dedicated models specific to transliterated data can only
deal with one script. Therefore, a natural research question
is: \emph{Can one make full use of an mPLM and a transliteration tool to build a strong baseline well-suited for transliterated data, \textbf{without any training}?}

To this end, this work presents a simple yet effective
framework: \textbf{Trans}literate-\textbf{M}erge-\textbf{I}nitialize
(\textbf{\textsc{TransMI}}). \frameworkname has three
stages. In the first stage, a
%rule-based
transliteration tool is used to transliterate all the subwords in the vocabulary of an mPLM into a common script (Latin in our case). Next, we merge the new subwords obtained in the previous step into the original tokenizer, where we propose three different modes to account for the problem of \emph{transliteration ambiguity} (different subwords in the vocabulary have the same transliteration). Lastly, we initialize the embeddings for the newly added subwords.
In this way, we modify the original mPLM so that it can deal
with transliterated data while not losing the ability to
process non-transliterated data. In contrast to the original
mPLM tokenizer that generates non-meaningful tokenization
for transliteration, the modified tokenizer generates tokens
that correspond well to natural linguistic units, as shown in Table \ref{tab:example_tokenization}.

% where we also have three modes corresponding to the modes proposed in the previous step. 

We validate \frameworkname by applying it to three recent
strong mPLMs that show remarkable crosslingual transfer and
evaluating the resulting models on a variety of downstream
tasks including sentence retrieval, text classification, and
sequence labeling. We evaluate each resulting model on
both transliterated and non-transliterated evaluation
datasets. We show that the models enhanced by our framework
not only achieve very similar performance on
non-transliterated data as their original mPLM counterparts
but also largely outperform them on transliterated data
across all downstream tasks.

The contributions are as follows: (i) We
present \frameworkname, a simple yet effective framework
that creates strong baselines from mPLMs for transliterated
data, without any training. (ii) We show that \frameworkname
boosts the performance on transliterated data
while not sacrificing performance on non-transliterated
data. (iii) We investigate in-depth how
different modes in the Merge (Initialize) step
in \frameworkname influence the performance. (iv) Through
fine-grained analysis, we show that \frameworkname benefits
languages from all script groups for transliterated data.

\section{Related Work}
% \subsection{Multilingual Pretrained Language Models}
% Models that are pretrained on a spectrum of languages with a specific self-learning objective, e.g., masked language modeling (MLM) \citep{devlin-etal-2019-bert} or causal language modeling \citep{radford2019language, brown2020gpt3} are referred to as mPLMs. From a structural point of view, there are three types of models: encoder-only models, encoder-decoder models, and decoder-only models. Typical encoder-only models, such as mBERT \citep{devlin-etal-2019-bert}, XLM-R \citep{conneau-etal-2020-unsupervised}, Glot500-m \citep{imanigooghari-etal-2023-glot500}, and XLM-V \citep{liang-etal-2023-xlm}, are usually good at natural language understanding tasks, e.g., text classification and sequential tagging. Encoder-decoder models, such as mBART \citep{liu-etal-2020-multilingual-denoising}, M2M-100 \citep{M2M2021fan} and mT5 \citep{xue-etal-2021-mt5}, usually achieve attractive performance on sequence-to-sequence tasks like machine translation. Decoder-only models, such as XGLM \citep{lin-etal-2022-shot}, mGPT \citep{shliazhko2022mgpt}, and BLOOM \citep{scao2022bloom}, are especially good at generation tasks. Recent scale-up to both the model size and the data size entitles these decoder-only models to surprising capability for various downstream tasks, and thus decoder-only models with a considerably large number of parameters are also referred to as large language models (LLMs) \citep{Chowdhery2023palm,touvron2023llama,achiam2023gpt}.

\subsection{Transliteration for Multilingual NLP}

Transliteration is the process of converting text from one
script into another \citep{chou1981conversion}. This process
does not involve translating meanings but rather represents
the source script symbols as faithfully as possible in the target
script. Transliteration has been proven to be an effective
method for improving neural machine translation between
languages that are written in different
scripts \citep{gheini2019universal,goyal-etal-2020-efficient,amrhein-sennrich-2020-romanization}. Transliteration
can also boost crosslingual alignment and transfer on a large scale when languages are transliterated into a common script,
especially for languages that are mutually influenced but
written in different
scripts \citep{dhamecha-etal-2021-role,muller-etal-2021-unseen,chau-smith-2021-specializing,purkayastha-etal-2023-romanization,moosa-etal-2023-transliteration,j-etal-2024-romansetu,ma2024exploring}. Recently, \citet{liu-etal-2024-translico} and \citet{xhelili-etal-2024-breaking} propose frameworks where
transliterations are used as an auxiliary input along with
the original-script text to improve the crosslingual
alignment across languages using different scripts.
Although this line of approaches improves crosslingual alignment \citep{liu2024transliterations},
extensive training for
adaptation to transliterated data is necessary.
In contrast, we propose
a simple framework to construct strong baselines
directly from existing mPLMs for transliterated data,
without any training.

\subsection{Vocabulary and Tokenizer Manipulation}
Training a new tokenizer on data of unseen languages and
optionally merging it with the original tokenizer is a
common way for efficient language
adaptation \citep{pfeiffer-etal-2021-unks,alabi-etal-2022-adapting,imanigooghari-etal-2023-glot500,liu-etal-2024-ofa}. Similarly,
adaptively manipulating the tokenizer and vocabulary also
shows strong performance improvement for domain-specific data
within the same
language \citep{sachidananda-etal-2021-efficient,lamproudis2022vocabulary,liu-etal-2023-task}. \citet{kajiura-etal-2023-vocabulary}
propose to replace certain subwords in a tokenizer with new
subwords learned from the domain-specific corpus for domain\
adaptation, thus not changing the vocabulary
size. Nevertheless, this line of work requires training to
learn good representations for the new subwords. Another
related work \citep{hofmann-etal-2022-embarrassingly},
instead of modifying the vocabulary, directly changes the
behavior of the tokenizer to preserve the morphological
structure, enhancing robustness and performance. Our
work also modifies the vocabulary and tokenizer by including
new subwords. In contrast to previous work, we initialize
the new subword embeddings by actively exploiting the
original mPLM embeddings. Thus the resulting model can be
directly adapted to the transliterated data without any
training.

\section{Preliminary: SentencePiece Unigram}
% The rest from HuggingFace, do change:

Unigram \citep{kudo-2018-subword} is a tokenization
algorithm for obtaining subword vocabulary, which is usually
used in conjunction with
SentencePiece \citep{kudo-richardson-2018-sentencepiece}. In
contrast to Byte-Pair Encoding
(BPE) \citep{gage1994new,sennrich-etal-2016-neural} or
WordPiece \citep{Schuster2012wordpiece,wu2016google},
Unigram is based on a language model that outputs multiple
subword segmentations with probabilities. In addition,
Unigram does not learn subwords through merging frequent
character combinations gradually as done by BPE. Instead, it
initializes a large number of units as its vocabulary and
progressively removes units that have low contributions to
the likelihood of the training corpus, until a
pre-defined vocabulary size is obtained. The optimization is
done by expectation-maximization (EM)
algorithm \citep{dempster1977maximum} and the overall
training objective is to maximize the marginal likelihood
$\mathcal{L}$:
\vspace{-0.1cm}
\begin{equation*}
\mathcal{L} =  \sum_{i=1}^{|\mathcal{D}|}\log \, P(X_{i}) = \sum_{i=1}^{|\mathcal{D}|}\log \,(\sum_{\boldsymbol{x} \in S(X_i)}P(\boldsymbol{x}))
\end{equation*}
where $\mathcal{D}$ is the training corpus, $X_i$ is the $i$th sentence in $\mathcal{D}$, and $S(X_i)$ is the set of all possible segmentation candidates for the input sentence $X_i$.

Once the Unigram tokenizer is trained, in addition to its vocabulary $V$, the model will also save a score, i.e., the log probability, learned from the training corpus, for each subword $w$ in $V$, as shown in Figure \ref{fig:framework}. This makes it possible for the model to provide the probability of each possible tokenization for a given sentence after training. In practice, the tokenizer is usually set to generate 
the most probable segmentation, i.e., the sequence of subwords that maximize the log probability:
\vspace{-0.1cm}
\begin{equation*}
    \boldsymbol{x}^* = \mbox{argmax}_{ \boldsymbol{x} \in S(X)} \sum_{w \in \boldsymbol{x}} \log P(w)
\end{equation*}
where $\boldsymbol{x}^*$ is the optimal tokenization given
 the sentence $X$ and $\log P(w)$ is the log probability of
 subword $w$.

% i removed this because the log probability is
% fixed, i.e., independent of the tokenziation x?
%in the tokenization $\boldsymbol{x}$.

\section{Methodology}
\begin{figure*}
    \setlength{\belowcaptionskip}{-0.2cm}
  \centering
  \includegraphics[width=\textwidth]{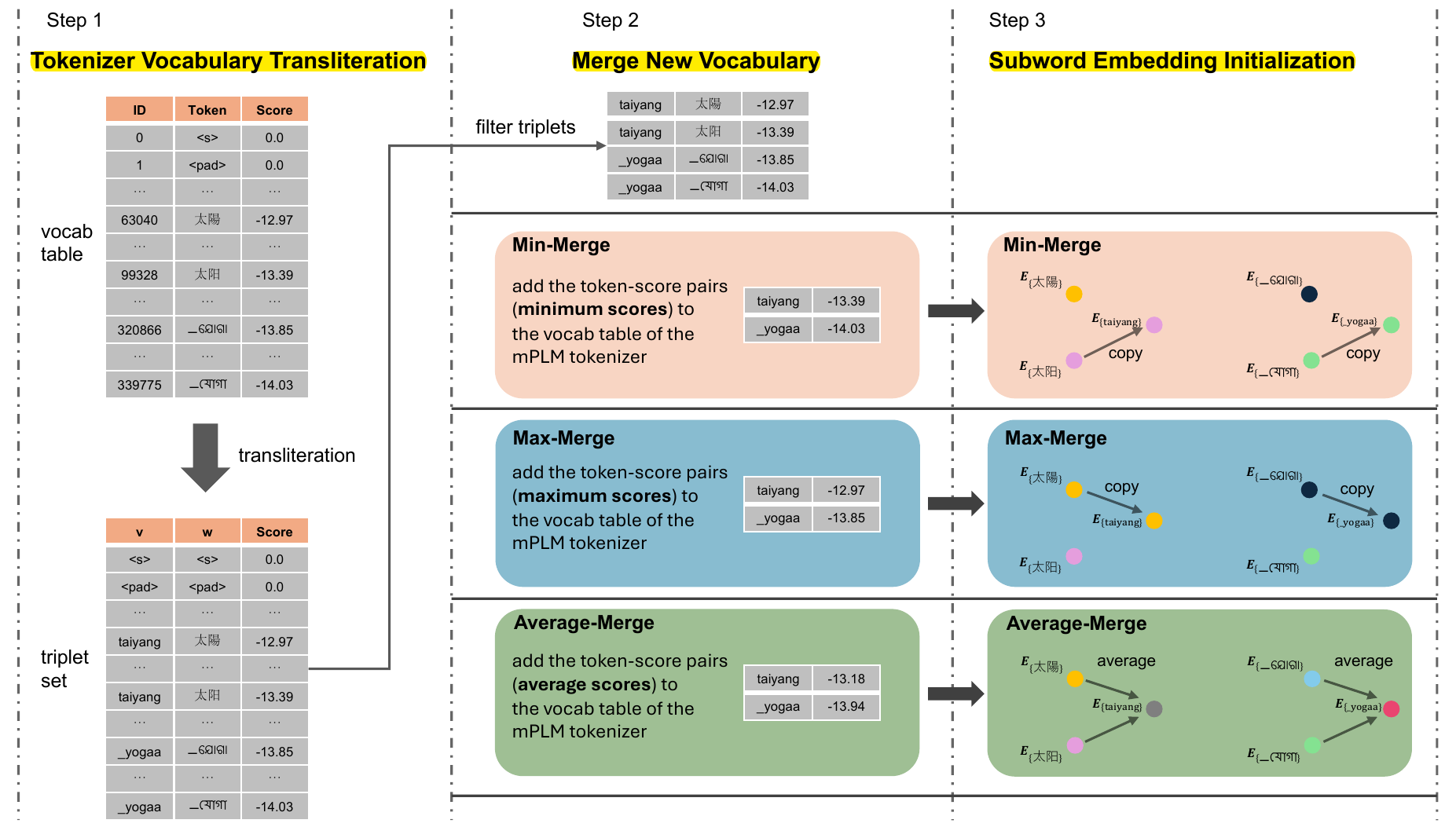}
  \caption{Overview of \textbf{\frameworkname}. We transliterate all the subwords from the vocabulary of the source mPLM tokenizer into Latin script in \textbf{Step 1}. We then merge the filtered triplets (ambiguous transliterations) into the tokenizer vocabulary table using one of the three proposed modes in \textbf{Step 2}. Note that we perform direct merge operations for the rest of the triplets that are not ambiguous (not shown in the figure). Lastly in \textbf{Step 3}, we initialize the embeddings for the newly added subwords according to the merge mode used in the previous step.}
  \label{fig:framework}
\end{figure*}

We introduce \frameworkname, a framework that makes full use of an existing mPLM and a transliteration tool to create strong baselines for transliterated data without any training.\footnote{In this work, we consider one special type of transliteration that involves converting non-Latin scripts into Latin script. This is also referred to as romanization.} There are three stages in \frameworkname: (1) transliterate the subwords in the original vocabulary; (2) merge the transliterated subwords into the original tokenizer; and (3) initialize the embeddings for the new subwords. In each stage, the information and knowledge from the mPLM and its tokenizer are carefully exploited. We illustrate the whole pipeline in Figure \ref{fig:framework} and introduce each stage in detail in the following. 

% unigram sentencepiece tokenizer
% https://huggingface.co/docs/transformers/en/tokenizer_summary#unigram

\subsection{Tokenizer Vocabulary Transliteration}
% TODO: give concrete examples of the vocabulary transliteration; 
The vocabulary of a multilingual tokenizer contains subwords
that are learned using tokenization algorithms such as
SentencePiece\footnote{The tokenizers used in this study are
all SentencePiece Unigram models where each subword is
associated with a log
probability. We refer to this log probability as the score.} \citep{kudo-richardson-2018-sentencepiece} on
a concatenation of data from different
languages \citep{conneau-etal-2020-unsupervised,imanigooghari-etal-2023-glot500}. As
a result, many subwords are in non-Latin script. An
intuitive way of adapting the vocabulary to Latin-script
transliterated data is to add the transliterations of these
non-Latin subwords into the vocabulary. Let
$V^{\text{orig}}$ be the vocabulary of the mPLM, and
$\texttt{Transli}$ be a deterministic transliterator. We create
a set of \emph{transliteration triplets} by applying
$\texttt{Transli}$ to every subword $w$ and associating it with
a score $s$, i.e., $w$'s log probability:
% V_{\text{trans}}
\begin{equation*}
T = \{(v, w, s) | v = \texttt{Transli}(w) \wedge w \in V^{\text{orig}} \}
\end{equation*}
It is important to note that $|T| = |V^{\text{orig}}|$ and there will be no duplicates in $T$. That is, given two elements: $(v_i, w_i, s_i)$ and $(v_j, w_j, s_j)$ where $v_i = v_j$, we always have $w_i \neq w_j$ (usually also $s_i \neq s_j$). For example, ``taiyang'' is the Latin transliteration for both
``\begin{CJK*}{UTF8}{gbsn}太阳''\end{CJK*} (``sun'' in simplified Chinese) and
``\begin{CJK*}{UTF8}{bsmi}太陽\end{CJK*}'' (``sun'' in
traditional Chinese).
%by \texttt{Uroman} \citep{hermjakob-etal-2018-box}. 
The score of ``\begin{CJK*}{UTF8}{bsmi}太陽\end{CJK*}'' is
higher than ``\begin{CJK*}{UTF8}{gbsn}太阳''\end{CJK*} since
``\begin{CJK*}{UTF8}{bsmi}太陽''\end{CJK*} is more frequent
and therefore has a higher log probability. 
Note the higher score only indicates the more frequent occurrence in the pretraining data of the mPLM but does not reflect the true population of people who write in that certain script.

% maybe it is important to show the distribution of how many one-to-one, and how many not one-to-one for different tokenizers. 

% TODO: talk about the difficulty because a new subword may be transliterated from multiple subwords, since transliteration is not a simple one-to-one mapping.

\subsection{Merge New Vocabulary}
Once we obtain $T$, we can modify the original vocabulary
$V^{\text{orig}}$ by adding the subword transliterations
$v_i$. However, we need to consider the possible
transliteration ambiguity while merging. For subword
transliterations that already exist in $V^{\text{orig}}$,
nothing needs to be done
-- this applies to all
subwords in Latin script without any
diacritics. For each subword transliteration $v'$ that has
a one-to-one relation to a $w'$, i.e.,
$\forall (v, w, s) \in T: v=v'\Rightarrow w=w'$,
we can simply add it with its
associated score to the vocabulary table.

For each of the remaining subwords $v'$,
we first define a new set of triplets
$U(v') := \{ (v,w,s) \in T | v=v'\}$.
Then,  to address the transliteration
ambiguity, we propose three different modes
% \footnote{Each
% mode is exclusive to the other: only one merge mode is
% selected each time when we merge the vocabulary.}
to merge
them into the vocabulary.

% \enote{hs}{i don't understand why we need the footnote -- i
% just find it confusing}
% \enote{yl}{i wanted to remind the readers; but yes we can remove it.}

\paragraph{Min-Merge Mode} In this mode, we select
the triplet whose score $s'$ is the \textbf{lowest} for the 
transliteration $v'$:
$$s'\dnrm{min}(v') = \min_{(v,w,s) \in U(v')}s$$
This mode will be favorable to preserve the less frequent
subwords. Adding the subword transliteration
$v'$ and its associated score $s'\dnrm{min}(v')$ to the vocabulary
table is likely to  alter the tokenizer's behavior negatively.
As a consequence, we
expect this mode will perform worst
among all modes.

%$$\{ (v_i, w_i, s_i) \mid \forall j, (v_i = v_j) \wedge i \neq j \Rightarrow (s_i < s_j) \}$$

\paragraph{Max-Merge Mode}
In contrast to Min-Merge Mode,
this mode selects the triplet
$(v', w', s') \in T$ whose score is the \textbf{highest}:
$$s'\dnrm{max}(v') = \max_{(v,w,s) \in U(v')}s$$
For high-frequency subwords,
this
mode replicates the previous tokenization behavior
for the original script: after transliteration,
we are likely to obtain the same tokenization.
Therefore, we expect this mode will achieve the
best performance.

%Then we use score $s'$ for $v'$:

\paragraph{Average-Merge Mode}
This mode \textbf{averages} the scores of all triplets containing $v'$.
%Let $U = \{ (v,w,s) \in T | v=v'\}$ and
$$ s'\dnrm{avg}(v') = \frac{\sum_{(v,w,s) \in U(v')}s}{|U(v')|}$$
We then add
subword $v'$ with score $ s'\dnrm{avg}(v')$ to the vocabulary.
Average-Merge Mode brings amount a change in tokenizer behavior that lies between Min-Merge Mode and Max-Merge Mode.\footnote{As is common in vocabulary extension and tokenizer merging
\citep{imanigooghari-etal-2023-glot500,lin2024mala}, we do
not renormalize the modified scores (to ensure the new
unigram distribution is a proper probability distribution)
because the tokenization behavior is only determined by the
order of scores.}

\subsection{Subword Embedding Initialization}
The last stage deals with embedding initialization for the
newly introduced subwords, which are transliterations of the subwords in the original vocabulary. 
Therefore, we aim to make full use of the
knowledge encoded in the original subword embedding matrix
$\boldsymbol{E}^{\text{orig}}$ to avoid any sort of
training. To achieve this, we create an additional embedding
matrix $\boldsymbol{E}^{\text{add}}$ for the new subwords
and initialize the embedding for each subword based on the
\textbf{correspondence} we obtain in the previous vocabulary merge
stage. Specifically, we directly copy the original embedding
for those new subwords that have a one-to-one
transliteration relation. 
For the rest of the subwords, we
initialize their embeddings according to which mode is used
in the last stage; this makes the resulting embeddings
consistent with
the updated tokenizer behavior.

\paragraph{Min-Merge Mode}
In Min-Merge Mode, we selected
the triple $(v',s'\dnrm{min}(v'),w')$.
Correspondingly, 
we initialize the embedding of $v'$ as
$w'$: $\boldsymbol{E}^{\text{add}}_{v'}
= \boldsymbol{E}^{\text{orig}}_{w'}$.

%That is, we
%initialize the embedding of $v'$ as the embedding of the
%subword $w'$ whose transliteration is $v'$ and score $s_i$
%is the \textbf{least} among all subwords in
%$V^{\text{orig}}$ that have $v_i$ as their transliterations.

\paragraph{Min-Merge Mode}
In Max-Merge Mode, we selected
the triple $(v',s'\dnrm{max}(v'),w')$.
Correspondingly, 
we initialize the embedding of $v'$ as
$w'$: $\boldsymbol{E}^{\text{add}}_{v'}
= \boldsymbol{E}^{\text{orig}}_{w'}$.

%\paragraph{Max-Merge Mode} Similarly, we use the triplets obtained in the last stage and initialize the embedding of a new subword $v_i$ as the embedding of the subword $w_i$ whose transliteration is $v_i$ and score $s_i$ is the \textbf{largest} among all subwords in $V^{\text{orig}}$ that have $v_i$ as their transliterations: $\boldsymbol{E}^{\text{add}}_{v_i} = \boldsymbol{E}^{\text{orig}}_{w_i}$.

\paragraph{Average-Merge Mode}
In Average-Merge Mode, we
averaged the scores of all $w'$ that are mapped to $v'$,
i.e., we averaged the scores in $U(v')$.
Correspondingly, 
we initialize the embedding of $v'$ as
the average of the $w'$:
$$\boldsymbol{E}^{\text{add}}_{v'}= \frac{\sum_{(v,w,s) \in U(v')}\boldsymbol{E}^{\text{orig}}_{w}}{|U(v')|}$$.

%$w'$: $\boldsymbol{E}^{\text{add}}_{v'}
%= \boldsymbol{E}^{\text{orig}}_{w'}$.

%We initialize the embedding of $v_i$ as the average embedding of all subwords with $v_i$ as their transliterations.

By choosing any mode, each embedding in
$\boldsymbol{E}^{\text{add}}$ is carefully initialized, and
in the same representation space as
$\boldsymbol{E}^{\text{orig}}$. To construct the final
embeddings, we simply concatenate
$\boldsymbol{E}^{\text{orig}}$ and
$\boldsymbol{E}^{\text{add}}$ and ensure the tokenizer
subword indices and their indices in the embeddings are
consistent.

\begin{table}
\footnotesize
\centering
\setlength{\tabcolsep}{2mm}{}
\begin{tabular}{lrrrrr}
\toprule
& \textbf{1} & \textbf{2} & \textbf{3} & \textbf{>3} & Total \\
\midrule
XLM-R & 97,456 & 6,866 & 1,380 & 1,088 & 106,790 \\
Glot500 & 123,001 & 8,373 & 1,706 & 1,274 & 134,354 \\
\bottomrule
% \bottomrule
\end{tabular}
\caption{Transliteration ambiguity of the newly added subwords (transliterations). For example, 97,456 indicates that, out of the total newly added 106,790 subwords for the XLM-R model, 97,456 subwords have a 1-to-1 relationship, i.e., the subword is the Latin transliteration of only 1 subword in the original XLM-R vocabulary. Most of the new subwords are not ambiguous.
}\label{tab:ambiguity}
\end{table}

\section{Experiments}
\subsection{Setups}

\begin{table*}[ht]
    % \setlength{\abovecaptionskip}{0cm}
    % \setlength{\belowcaptionskip}{-0.4cm}
    % \scriptsize
    \footnotesize
    \centering
    \setlength{\tabcolsep}{0.76mm}{}
    \begin{tabular}{lrrrrrrrrrrrrrrrrrr}
        \toprule
        & \multicolumn{3}{c}{SR-B} & \multicolumn{3}{c}{SR-T} & \multicolumn{3}{c}{Taxi1500} &  \multicolumn{3}{c}{SIB200} & \multicolumn{3}{c}{NER} & \multicolumn{3}{c}{POS}\\
        \cmidrule(lr){2-4} \cmidrule(lr){5-7} \cmidrule(lr){8-10} \cmidrule(lr){11-13} \cmidrule(lr){14-16} \cmidrule(lr){17-19}
        & tail & head & all & tail & head & all & tail & head & all & tail & head & all & tail & head & all & tail & head & all\\
        \midrule
XLM-R & 7.0 & 28.6 & 12.5 & 26.5 & 35.4 & 32.9 & 11.4 & 34.8 & 17.4 & 43.0 & 52.7 & 47.4 & 46.3 & 46.2 & 46.3 & 34.1 & 59.0 & 51.3 \\
XLM-R (Max-Merge)  & \textbf{7.4} & \textbf{35.8} & \textbf{14.6} & \textbf{30.1} & \textbf{48.2} & \textbf{43.0} & \textbf{13.6} & \textbf{47.0} & \textbf{22.1} & \textbf{48.3} & \textbf{73.0} & \textbf{59.5} & \textbf{46.7} & \textbf{53.7} & \textbf{50.5} & \textbf{37.8} & \textbf{70.1} & \textbf{60.2} \\
\midrule
Glot500 & 33.1 & 31.6 & 32.7 & 44.9 & 42.3 & 43.0 & 41.5 & 36.4 & 40.2 & 59.3 & 56.2 & 57.9 & 54.0 & 49.0 & 51.3 & 48.9 & 59.8 & 56.4 \\
Glot500 (Max-Merge) & \textbf{34.3} & \textbf{38.4} & \textbf{35.4} & \textbf{49.2} & \textbf{55.5} & \textbf{53.7} & \textbf{45.1} & \textbf{48.9} & \textbf{46.0} & \textbf{66.8} & \textbf{74.7} & \textbf{70.4} & \textbf{57.3} & \textbf{57.2} & \textbf{57.2} & \textbf{52.5} & \textbf{68.8} & \textbf{63.8} \\
\midrule
\textsc{Furina} & 47.9 & 51.2 & 48.7 & 55.4 & 53.6 & 54.1 & 45.0 & 43.0 & 44.5 & 60.7 & 59.9 & 60.3 & 54.4 & 52.2 & 53.2 & 54.3 & 67.4 & 63.4 \\
\textsc{Furina} (Max-Merge)& \textbf{49.4} & \textbf{54.2} & \textbf{50.6} & \textbf{56.4} & \textbf{59.1} & \textbf{58.3} & \textbf{49.6} & \textbf{53.0} & \textbf{50.5} & \textbf{66.7} & \textbf{74.1} & \textbf{70.1} & \textbf{57.6} & \textbf{58.5} & \textbf{58.1} & \textbf{55.9} & \textbf{71.7} & \textbf{66.8} \\
        \bottomrule
    \end{tabular}
    \caption{Performance of three model types on \textbf{transliterated} evaluation datasets across 5 random seeds. We report the performance as an average over head, tail, and all language-scripts for each model variant. Max-merge models consistently outperform the original model on transliterated evaluation data.
    \textbf{Bold}: best result per model type.}
    \label{tab:results_transliterated}
\end{table*}

\begin{table*}[ht]
    % \setlength{\abovecaptionskip}{0cm}
    % \setlength{\belowcaptionskip}{-0.4cm}
    % \scriptsize
    \footnotesize
    \centering
    \setlength{\tabcolsep}{0.76mm}{}
    \begin{tabular}{lrrrrrrrrrrrrrrrrrr}
        \toprule
        & \multicolumn{3}{c}{SR-B} & \multicolumn{3}{c}{SR-T} & \multicolumn{3}{c}{Taxi1500} &  \multicolumn{3}{c}{SIB200} & \multicolumn{3}{c}{NER} & \multicolumn{3}{c}{POS}\\
        \cmidrule(lr){2-4} \cmidrule(lr){5-7} \cmidrule(lr){8-10} \cmidrule(lr){11-13} \cmidrule(lr){14-16} \cmidrule(lr){17-19}
        & tail & head & all & tail & head & all & tail & head & all & tail & head & all & tail & head & all & tail & head & all\\
        % \midrule
        % XLM-V & 10.1 & \underline{60.3} & 22.9 & 39.3 & \textbf{78.2} & \textbf{67.1} & 17.7 & 61.1 & 28.7 & 53.2 & 65.3 & 59.8 & 45.4 & \textbf{76.6} & 67.0 \\
        \midrule
XLM-R & \textbf{7.4} & \textbf{54.2} & \textbf{19.3} & 32.6 & \textbf{66.2} & \textbf{56.6} & \textbf{13.5} & \textbf{58.7} & \textbf{25.0} & \textbf{49.5} & \textbf{81.1} & \textbf{63.9} & \textbf{47.6} & \textbf{61.0} & \textbf{54.9} & \textbf{42.7} & 76.4 & 66.0 \\
XLM-R (Max-Merge) & \textbf{7.4} & 53.3 & 19.1 & \textbf{33.1} & 65.1 & 56.0 & 13.2 & 58.0 & 24.6 & 46.9 & 80.7 & 62.2 & 46.6 & 60.7 & 54.3 & \textbf{42.7} & \textbf{76.5} & \textbf{66.1} \\
\midrule
Glot500 & \textbf{43.2} & \textbf{59.0} & \textbf{47.3} & \textbf{59.8} & \textbf{75.0} & \textbf{70.7} & \textbf{52.5} & \textbf{60.9} & \textbf{54.6} & 68.5 & 80.4 & 73.9 & \textbf{60.8} & \textbf{63.7} & \textbf{62.4} & 62.0 & \textbf{76.0} & \textbf{71.7} \\
Glot500 (Max-Merge) & 41.7 & 57.8 & 45.8 & 58.3 & 72.8 & 68.7 & 51.5 & 60.8 & 53.8 & \textbf{69.5} & \textbf{81.2} & \textbf{74.8} & 59.6 & 63.2 & 61.6 & \textbf{62.1} & \textbf{76.0} & \textbf{71.7} \\
\midrule
\textsc{Furina} & \textbf{55.3} & \textbf{66.2} & \textbf{58.1} & \textbf{62.1} & \textbf{71.5} & \textbf{68.8} & 55.9 & \textbf{63.8} & 57.9 & 70.3 & 82.2 & 75.7 & \textbf{60.5} & 63.9 & \textbf{62.4} & \textbf{63.3} & 75.7 & 71.9 \\
\textsc{Furina} (Max-Merge) & \textbf{55.3} & 65.9 & 58.0 & 60.9 & 70.6 & 67.9 & \textbf{56.6} & \textbf{63.8} & \textbf{58.4} & \textbf{71.8} & \textbf{82.5} & \textbf{76.7} & 60.1 & \textbf{64.2} & \textbf{62.4} & 62.1 & \textbf{76.5} & \textbf{72.1} \\
        \bottomrule
    \end{tabular}
    \caption{Performance of three model types on \textbf{non-transliterated} evaluation datasets across 5 random seeds. We report the performance as an average over head, tail, and all language-scripts for each model variant. Max-merge models perform close to the original models.
    \textbf{Bold}: best result per model type.}
    \label{tab:results_non_transliterated}
\end{table*}

We apply the proposed framework \frameworkname to three strong mPLMs: XLM-R, Glot500, and \textsc{Furina}. XLM-R \citep{conneau-etal-2020-unsupervised} is pretrained on 100 languages using masked language modeling (MLM) \citep{devlin-etal-2019-bert}. Glot500 \citep{imanigooghari-etal-2023-glot500} is a continued pretrained model from XLM-R on Glot500-c dataset that covers more than 500 languages. \textsc{Furina} \citep{liu-etal-2024-translico} is a post-aligned version of Glot500, which is fine-tuned using 5\% of pretraining data of Glot500.\footnote{The data is transliterated into Latin script and both the transliterated and the original data are used in fine-tuning.} We use \texttt{Uroman} \citep{hermjakob-etal-2018-box} as the rule-based transliteration tool. Note that the tokenizers of Glot500 and \textsc{Furina} are the same. We show the number of newly added subwords and the transliteration ambiguity in Table \ref{tab:ambiguity}. We use the \textbf{base} version of each model (their architectures are the same) for a fair comparison. There are 9 resulting models (3 merge modes $\times$ 3 model types) in total. When evaluating the models, we use both the non-transliterated evaluation datasets (the original ones) and the transliterated Latin-script evaluation datasets, which are obtained by transliterating the original evaluation datasets using \texttt{Uroman}. Following \citet{imanigooghari-etal-2023-glot500}, we refer to language-scripts supported by XLM-R as the \textbf{head} languages and the remaining language-scripts -- those that are supported by Glot500 as the \textbf{tail} languages.\footnote{A language-script is a combination of its ISO 639-3 and script codes.}

\subsection{Downstream Tasks}

We consider the following three evaluation types. For each
type, we consider two evaluation datasets. The evaluation is
performed in an English-centric crosslingual zero-shot
fashion: fine-tuning on the English train set, selecting the
best checkpoint on the English development set, and then evaluating the best checkpoint on the test sets of all other language-scripts.
An exception is Sentence Retrieval in that it does not involve any fine-tuning. For all tasks, only the subset of languages (head and tail languages) supported by Glot500 are considered. Details of the used dataset and hyperparameter settings for fine-tuning are reported in \secref{hyperparam}.

\paragraph{Sentence Retrieval.} We use Bible (SR-B) and Tatoeba \citep{artetxe-schwenk-2019-massively} (SR-T). The similarity is calculated using the mean pooling of contextualized word embeddings at the 8th layer. 

\paragraph{Text Classification.} We use Taxi1500 \citep{ma2023taxi1500} and SIB200 \citep{adelani-etal-2024-sib}.

\paragraph{Sequence Labeling.} We use WikiANN for named entity recognition (NER) \citep{pan-etal-2017-cross} and Universal Dependencies \citep{de-marneffe-etal-2021-universal} for Part-Of-Speech (POS) tagging.

% TODO general results using aggregated
% script using radar graph

% the reason why the performance after modification to the tokenizer for the non-Latn data is because the tokenization for the source language English is changed.

\subsection{Results and Discussion}

There are two goals to achieve with \frameworkname: (1) we want to build strong baselines directly from existing mPLMs for transliterated data, and (2) we want to preserve the mPLMs' ability to deal with the original scripts, i.e., non-transliterated data. 
Therefore, we evaluate the original mPLMs and their corresponding variants modified by \frameworkname on both the transliterated evaluation datasets (all scripts are converted into Latin script) and the non-transliterated (original) evaluation datasets. We notice that the different merge modes offer very similar performance while the \textbf{Max-Merge} mode slightly outperforms the other modes. As a result, we only show the performance of the \textbf{Max-Merge} mode in this section (Table \ref{tab:results_transliterated} and \ref{tab:results_non_transliterated}). The comparison between different modes is presented and discussed in \secref{mode}.

\begin{table*}[ht]
    \setlength{\belowcaptionskip}{-0.2cm}
    \scriptsize
    % \footnotesize
    \centering
    \setlength{\tabcolsep}{1.4mm}{}
    \begin{tabular}{lrrrrrrrrrrrrrrrrrr}
        \toprule
        & \multicolumn{3}{c}{SR-B} & \multicolumn{3}{c}{SR-T} & \multicolumn{3}{c}{Taxi1500} &  \multicolumn{3}{c}{SIB200} & \multicolumn{3}{c}{NER} & \multicolumn{3}{c}{POS}\\
        \cmidrule(lr){2-4} \cmidrule(lr){5-7} \cmidrule(lr){8-10} \cmidrule(lr){11-13} \cmidrule(lr){14-16} \cmidrule(lr){17-19}
        & tail & head & all & tail & head & all & tail & head & all & tail & head & all & tail & head & all & tail & head & all\\
        % \midrule
        % XLM-V & 10.1 & \underline{60.3} & 22.9 & 39.3 & \textbf{78.2} & \textbf{67.1} & 17.7 & 61.1 & 28.7 & 53.2 & 65.3 & 59.8 & 45.4 & \textbf{76.6} & 67.0 \\
        \midrule
XLM-R (Min-Merge) & \textbf{7.4} & \underline{34.7} & \underline{14.4} & 28.7 & \underline{46.4} & \underline{41.3} & \textbf{14.4} & 45.0 & \underline{22.1} & \textbf{49.6} & \textbf{73.7} & \textbf{60.5} & 45.9 & 52.3 & 49.3 & \textbf{38.0} & 69.1 & 59.5 \\
XLM-R (Average-Merge) & \textbf{7.4} & 34.1 & 14.2 & \underline{29.0} & 45.4 & 40.7 & \underline{14.3} & \underline{46.2} & \textbf{22.4} & {47.1} & 70.7 & {57.8} & \textbf{47.2} & \underline{53.5} & \textbf{50.6} & {37.6} & \underline{69.4} & \underline{59.6} \\
XLM-R (Max-Merge) & \textbf{7.4} & \textbf{35.8} & \textbf{14.6} & \textbf{30.1} & \textbf{48.2} & \textbf{43.0} & 13.6 & \textbf{47.0} & \underline{22.1} & \underline{48.3} & \underline{73.0} & \underline{59.5} & \underline{46.7} & \textbf{53.7} & \underline{50.5} & \underline{37.8} & \textbf{70.1} & \textbf{60.2} \\
\midrule
Glot500 (Min-Merge) & 34.1 & 36.2 & 34.6 & 48.8 & 53.0 & 51.8 & \underline{46.1} & 48.0 & \underline{46.6} & \textbf{67.1} & 73.8 & \underline{70.1} & \textbf{57.4} & 55.8 & \underline{56.6} & \textbf{52.5} & 67.1 & 62.6 \\
Glot500 (Average-Merge) & \underline{34.2} & \underline{37.4} & \underline{35.0} & \textbf{49.4} & \underline{54.8} & \underline{53.2} & \textbf{47.2} & \textbf{49.8} & \textbf{47.9} & {66.6} & \underline{73.9} & {69.9} & {56.7} & \underline{56.1} & {56.4} & \underline{52.0} & \underline{68.3} & \underline{63.3} \\
Glot500 (Max-Merge) & \textbf{34.3} & \textbf{38.4} & \textbf{35.4} & \underline{49.2} & \textbf{55.5} & \textbf{53.7} & {45.1} & \underline{48.9} & {46.0} & \underline{66.8} & \textbf{74.7} & \textbf{70.4} & \underline{57.3} & \textbf{57.2} & \textbf{57.2} & \textbf{52.5} & \textbf{68.8} & \textbf{63.8} \\
\midrule
\textsc{Furina} (Min-Merge) & \underline{49.3} & 52.3 & 50.1 & \underline{56.2} & 58.0 & 57.4 & \underline{48.5} & 50.2 & 49.0 & \underline{66.7} & 72.8 & 69.5 & \textbf{58.2} & \textbf{59.0} & \textbf{58.6} & 55.5 & 71.0 & \underline{66.2} \\
\textsc{Furina} (Average-Merge) & \textbf{49.4} & \underline{53.5} & \underline{50.5} & \underline{56.2} & \underline{58.5} & \underline{57.8} & {48.4} & \underline{51.4} & \underline{49.2} & \textbf{67.6} & \textbf{75.1} & \textbf{71.0} & {57.0} & {57.7} & {57.4} & \textbf{56.1} & \underline{71.6} & \textbf{66.8} \\
\textsc{Furina} (Max-Merge) & \textbf{49.4} & \textbf{54.2} & \textbf{50.6} & \textbf{56.4} & \textbf{59.1} & \textbf{58.3} & \textbf{49.6} & \textbf{53.0} & \textbf{50.5} & \underline{66.7} & \underline{74.1} & \underline{70.1} & \underline{57.6} & \underline{58.5} & \underline{58.1} & \underline{55.9} & \textbf{71.7} & \textbf{66.8} \\
        \bottomrule
    \end{tabular}
    \caption{Performance of three merge modes applied to three mPLMs on \textbf{transliterated} evaluation datasets across 5 random seeds. The performance difference among the three modes are small but Min-Merge mode is favorable to tail languages while Max-Merge mode is favorable to head languages. In general, Max-Merge mode achieves the overall best performance. \textbf{Bold} (\underline{underlined}): best (second-best) result for each task in each model type.}
    \label{tab:results_modes}
\end{table*}

\subsubsection{Evaluation on Transliterated Data}

We report performance on transliterated data in Table \ref{tab:results_transliterated}. 
We observe consistent improvement across all language-scripts and tasks.
Generally, head languages enjoy a higher increase than tail languages.
In addition, sentence-level tasks seem to benefit more from the method than the token-level tasks.

The original mPLMs are not pretrained on transliterated data and thus they perform suboptimally on transliterated evaluation datasets. 
Modifying these mPLMs with \frameworkname equips these models with the ability to deal with the transliterated texts, as new subwords (transliterations of the subwords covered by the original mPLMs) are included and their embeddings are wisely initialized. 
Consequently, the resulting models can process the transliterated data and achieve good performance.

It can be observed that the improvement by modifying \textsc{Furina} is relatively smaller than on other model types. We hypothesize this is because \textsc{Furina} is already fine-tuned on transliterated data through MLM and transliteration contrastive modeling \citep{liu-etal-2024-translico}. In this way, even though \textsc{Furina}'s vocabulary is the same as Glot500, i.e., not extended and adapted accordingly for Latin transliterations, its fine-tuning phase still helps the model gain some knowledge beneficial for processing and understanding transliterated data.

\subsubsection{Evaluation on Non-transliterated Data}
% by table
We report performance on non-transliterated data in
Table \ref{tab:results_non_transliterated}. Across all
tasks, modified models achieve performance very close to the original mPLMs, with generally negligibly small performance degradation. This is expected as the vocabulary is only augmented with new subwords in Latin script (transliteration) and therefore the tokenization results for non-Latin texts remain the same (their embeddings are also not altered at all).

On the other hand, the slight decrease in performance is not surprising. The included new subwords will influence the tokenization results for all languages written in Latin script, including English. As the evaluation is done in an English-centric manner, even if the target language is not written in Latin script, altered English tokenization can still influence the crosslingual transfer performance. When the transfer target language is also written in Latin, the tokenization results and representation are changed on both the source side and target side, the performance therefore varies.

\section{Analysis}
\subsection{Which Merge Mode Wins}\seclabel{mode}

To explore how different merge modes (Min-Merge,
Average-Merge, and Max-Merge) influence the English-centric zero-shot crosslingual transfer, 
we report the performance of the three variants of
each model type in Table \ref{tab:results_modes}. Generally,
the performance differences among the three merge modes are
very small across model types and downstream tasks. This can be
explained by the fact that most of the newly added subwords
(transliterations of subwords in the original mPLM
vocabulary) are not ambiguous: 91\% of subwords in XLM-R and
92\% in Glot500 (also \textsc{Furina}) have 1-to-1
relations, as shown in Table \ref{tab:ambiguity}. Each merge
mode simply does the same copy operation for these
unambiguous subwords, and operates differently only on the
remaining ambiguous subwords (the transliteration
corresponds to more than one subword in the original
vocabulary), which is a relatively small portion.

Although the difference is small, we observe that the
Min-Merge mode supports tail languages better while the
Max-merge mode supports head languages better. We
hypothesize that the scores (log probabilities) of some
subwords from tail languages (usually low-resource
languages) are small, and the Min-Merge mode preserves the
small scores and tokenization behavior when tokenizing texts
that contain these tokens, therefore is favorable to tail
languages. Similarly, the scores for some subwords from
high-resource languages are large because of higher frequencies in the pretraining data, and the Max-Merge mode
keeps their priority in tokenization. In general, the
Max-Merge mode is the best option, as it has the
overall best performance across all tasks in each model type.

\begin{figure*}[htbp]
\centering
\subfigure[Sentence Retrieval]{
\begin{minipage}[t]{0.31\linewidth}
\centering
\includegraphics[width=\textwidth]{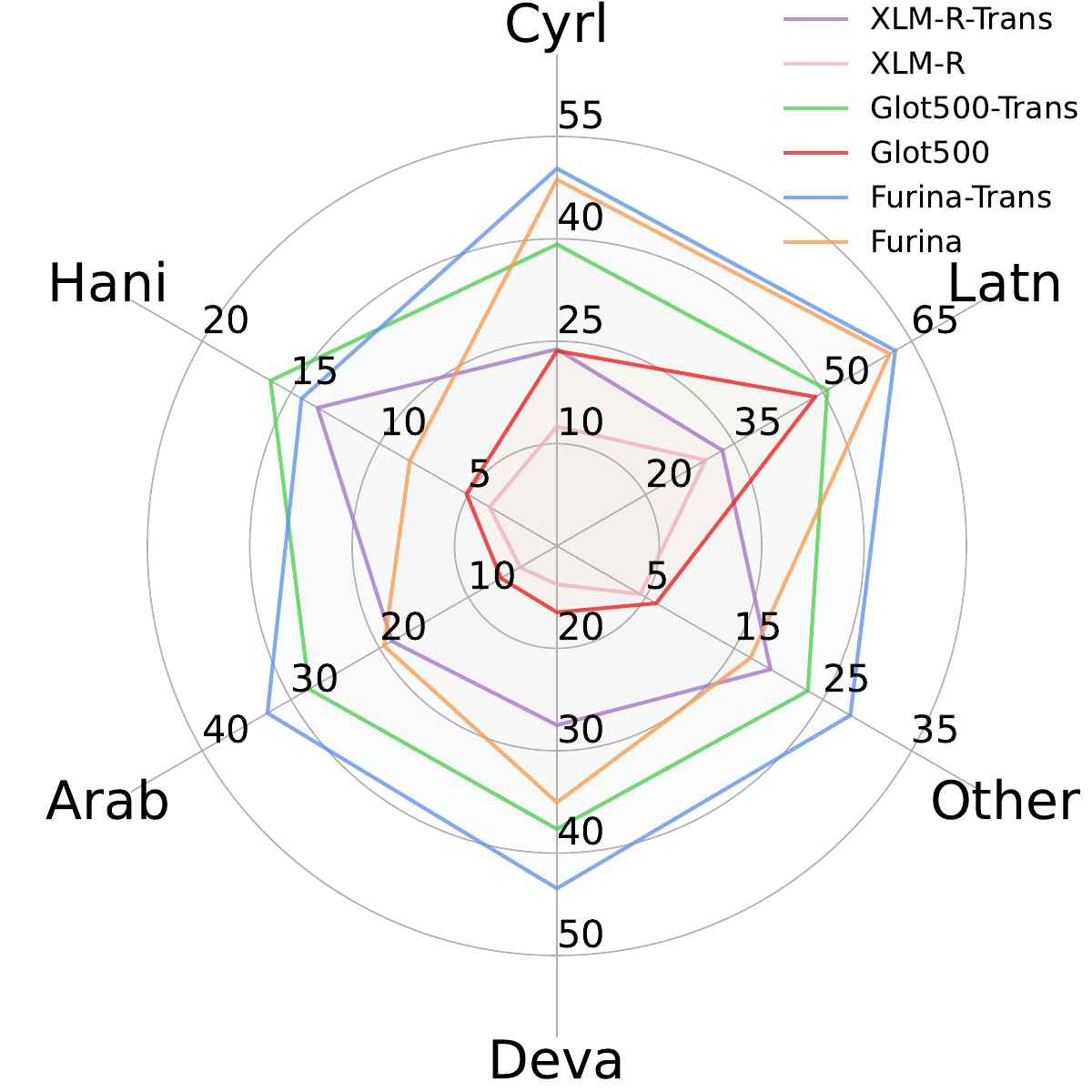}
\end{minipage}%
}%
\subfigure[Text Classification]{
\begin{minipage}[t]{0.31\linewidth}
\centering
\includegraphics[width=\textwidth]{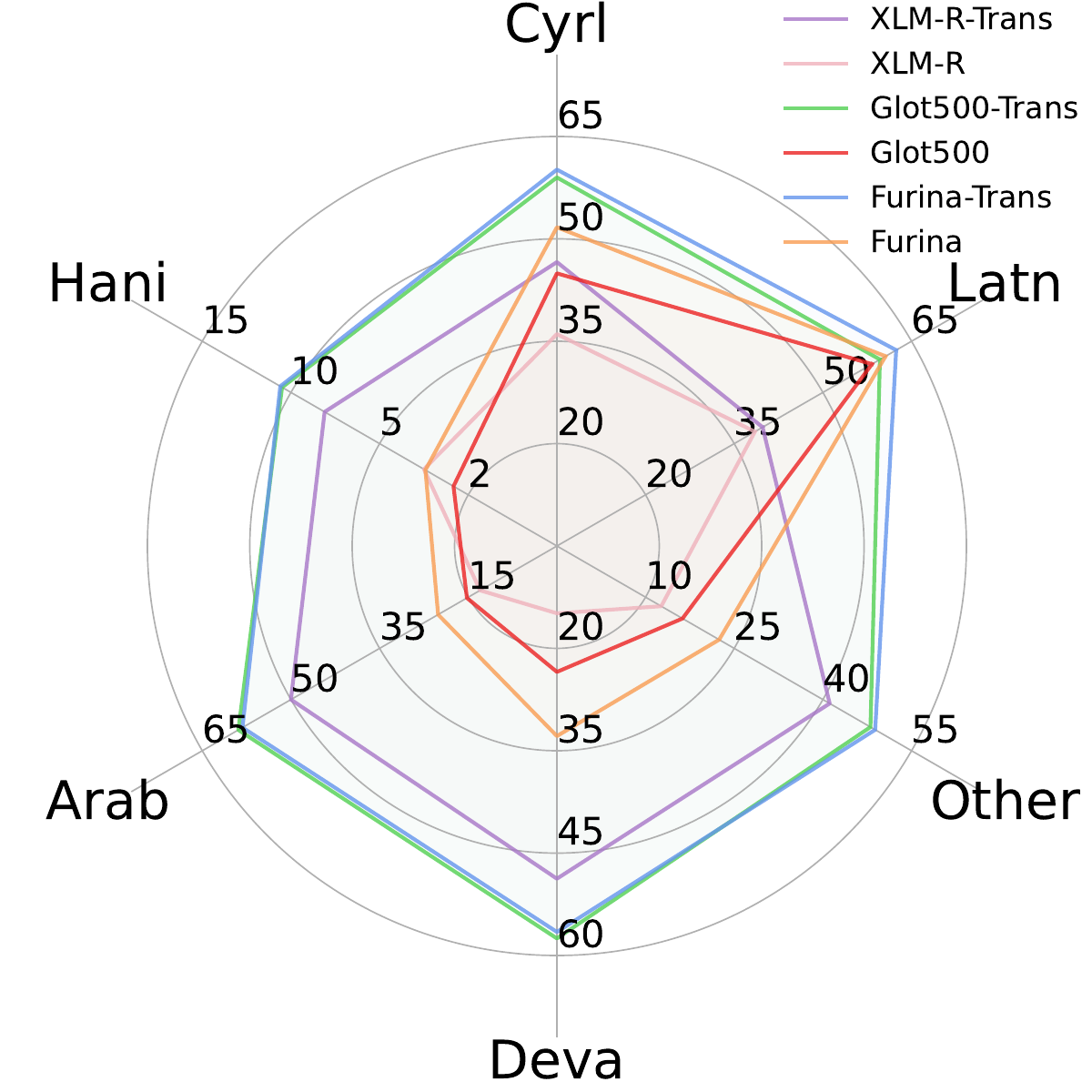}
\end{minipage}%
}%
\subfigure[Sequence Labeling]{
\begin{minipage}[t]{0.31\linewidth}
\centering
\includegraphics[width=\textwidth]{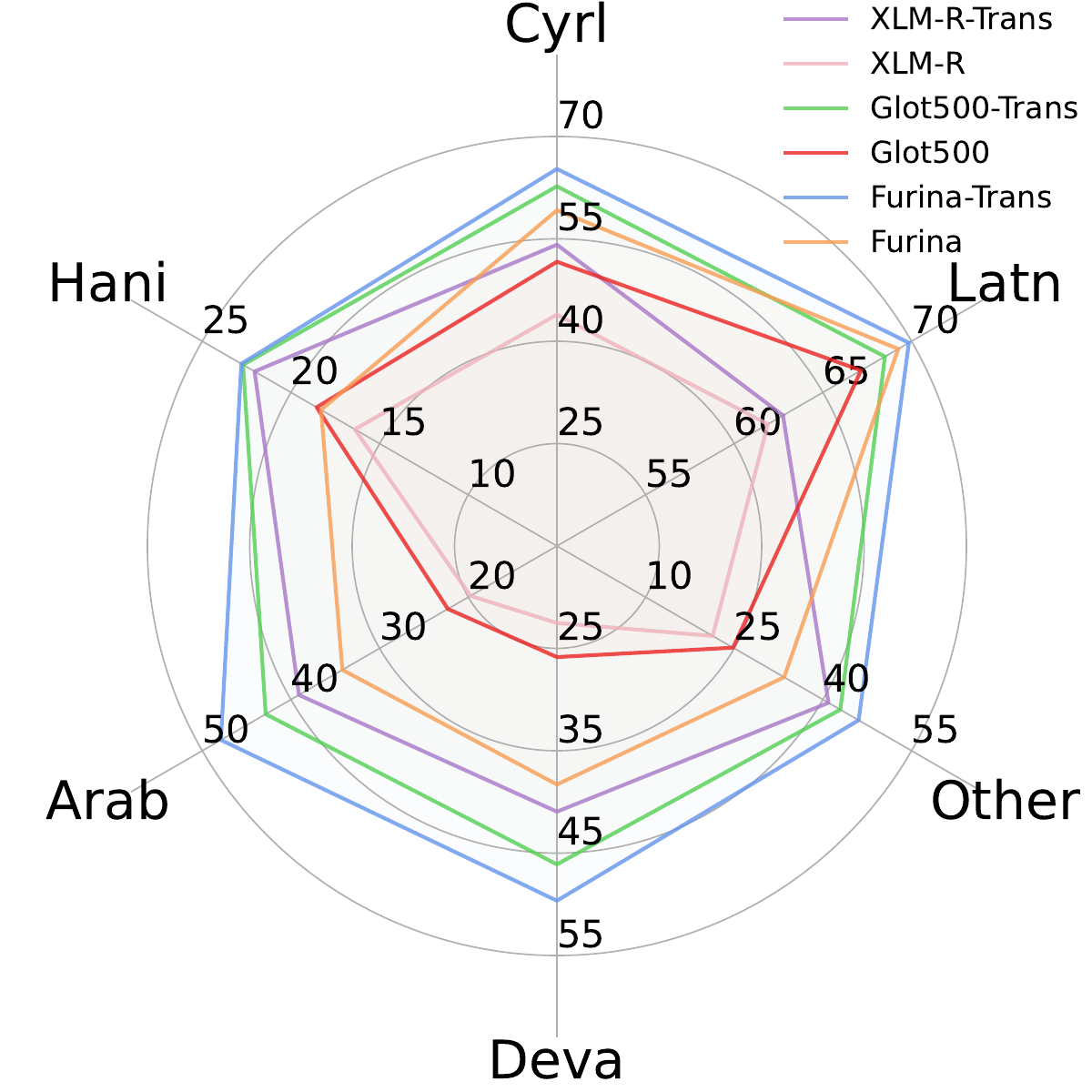}
\end{minipage}%
}
\caption{Qualitative comparison between the original mPLMs and \frameworkname (Max-Merge mode) models (denoted with ``-Trans'') on transliterated evaluation datasets. We compute the average performance for each evaluation type (e.g., Sentence Retrieval is the average of SR-B and SR-T) for different script groups. Each language is placed into a group according to the script that the language is originally written in. The script groups are: \textbf{Latn} (Latin), \textbf{Cyrl} (Cyrillic), \textbf{Hani} (Hani), \textbf{Arab} (Arabic), and \textbf{Deva} (Devanagari). Languages not written in these scripts are placed into \textbf{Other}. \frameworkname-modified models consistently outperform the original mPLMs across all tasks and groups.}
% especially for languages that are originally written in scripts other than Latn. 
\label{fig:radar}
\end{figure*}

\enote{hs}{does this have to be said:
``The stride of each axis in the chart is different.''? It is obvious?}
\enote{yl}{yes usually it's obvious, but some readers might just didn't pay enough attention.}

\subsection{Which Scripts Benefit}

It is also important to know whether and how much \frameworkname-modified models outperform the original mPLMs on languages that are originally written in different scripts in the transliterated evaluation.
Therefore, we compare the original mPLMs and their modified variants (using the Max-Merge mode) on the three evaluation types (transliterated evaluation) across different scripts in which the transfer target languages are originally written, as shown in Figure \ref{fig:radar}. Globally, each \frameworkname-modified model consistently achieves better performance than its counterpart across all script groups and all evaluation types, since the original mPLMs are not specifically trained on transliterated data, except for \textsc{Furina}, for which we observe similar performance occasionally. For example, \textsc{Furina} and \textsc{Furina}-Trans achieve very close performance in the Cyrl group in Sentence Retrieval. 

We observe that the smallest improvement comes from the Hani group and the Latn group. This is not surprising and can be explained by the fact that the former is logograms and transliterated words potentially lose semantic or contextual nuances and are more prone to ambiguity \citep{liu-etal-2024-translico} while the latter does not change much after \texttt{Uroman} transliteration (only diacritics are removed). The rest of the script groups, i.e., Cyrl, Arab, Deva and Other, all enjoy large improvements, which indicates that \frameworkname is effective in creating strong baselines for transliterated data for languages originally written in phonetic scripts.

% by radar graph
% Which script benefits the most by applying TransMI (of course, this is only with respect to transliterated data)
\subsection{Transliterated vs. Non-transliterated}

\begin{table}
\footnotesize
\centering
\setlength{\tabcolsep}{1.0mm}{}
\begin{tabular}{lrrrrrr}
\toprule
& Latn & Cyrl & Hani & Arab & Deva & Other \\
\midrule
\multirow{2}{*}{\bf Sentence Retrieval} & \textbf{64.6} & \textbf{69.0} & \textbf{43.1} & \textbf{58.2} & \textbf{68.8} & \textbf{55.2} \\
& 62.2 & 50.3 & 14.4 & 32.6 & 43.4 & 28.1 \\
\midrule
\multirow{2}{*}{\bf Text Classification} & \textbf{65.7} & \textbf{73.2} & \textbf{34.2} & \textbf{73.0} & \textbf{76.5} & \textbf{71.5} \\
& 62.4 & 60.2 & 10.6 & 58.2 & 56.6 & 48.8 \\
\midrule
\multirow{2}{*}{\bf Sequence Labeling} & \textbf{70.8} & \textbf{72.0} & \textbf{29.2} & \textbf{62.8} & \textbf{59.6} & \textbf{60.0} \\
& 69.8 & 65.2 & 22.8 & 47.9 & 49.6 & 46.0 \\
\bottomrule
% \bottomrule
\end{tabular}
\caption{Comparison between the performance of \textsc{Furina} (Max-Merge) on non-transliterated and transliterated evaluation datasets for different script groups. For each evaluation type, the results on non-transliterated (resp. transliterated) data are in the first (resp. second) row. \textsc{Furina} (Max-Merge) consistently performs better on non-transliterated data than transliterated data.
}\label{tab:trans_vs_non_perf}
\end{table}

\begin{table}
\setlength{\belowcaptionskip}{-0.4cm}
\footnotesize
\centering
\setlength{\tabcolsep}{0.7mm}{}
\begin{tabular}{lrrrrrr}
\toprule
& Latn & Cyrl & Hani & Arab & Deva & Other \\
\midrule
\multirow{2}{*}{\textsc{Furina}} & 45.7 & 39.4 & 42.7 & 42.3 & 44.1 & 44.5 \\
& 45.5 & 43.8 & 34.9 & 50.8 & 53.7 & 57.0 \\
\midrule
\multirow{2}{*}{\textsc{Furina} (Max-Merge)} & 44.8 & 39.4 & 42.7 & 42.3 & 44.1 & 44.5 \\
& \textbf{43.3} & \textbf{36.0} & \textbf{30.9} & \textbf{36.3} & \textbf{38.4} & \textbf{38.9} \\
\bottomrule
% \bottomrule
\end{tabular}
\caption{Average sequence length of SR-B dataset averaged by script group. The results on non-transliterated (resp. transliterated) data are in the first (resp. second) row. The tokenizer of \textsc{Furina} (Max-Merge) has consistently shorter sequence lengths on transliterated data compared with the original \textsc{Furina} tokenizer.
}\label{tab:trans_vs_non_token}
\end{table}

We also want to explore how the performance varies before and after the evaluation datasets are transliterated. 
Therefore, we present the comparison between transliterated evaluation and non-transliterated evaluation across different script groups, using \textsc{Furina} (Max-Merge) as a case study, in Table \ref{tab:trans_vs_non_perf}. Not surprisingly, the performance of all script groups drops after transliterating the evaluation datasets.
There are mainly two reasons: \textbf{(1)} the embeddings of the newly added subwords that have transliteration ambiguity are not suitable for each occurrence \textbf{within the same language}, e.g., ``shiwu'' is the transliteration for both
``\begin{CJK*}{UTF8}{gbsn}食物''\end{CJK*} (``food'' in Chinese) and
``\begin{CJK*}{UTF8}{gbsn}时务\end{CJK*}'' (``current
affairs'' in Chinese) and \textbf{across different
languages}, e.g., ``miso'' is the transliteration for
both \textgreek{mis'o} (``half'' in Greek) and
``\begin{CJK*}{UTF8}{gbsn}味噌\end{CJK*}'' (``soybean paste'' in
Japanese); \textbf{(2)}
the tokenization result is different before and after
transliterating a sentence into Latin script. Although our method tries to preserve the tokenization behavior, it is impossible to prevent changes completely. As shown in Table \ref{tab:trans_vs_non_token}, the average sequence length changes in all script groups for \textsc{Furina} (Max-Merge), even for the Latin group, since the newly added subwords inevitably also change the tokenization for Latin-script languages.
The different tokenizations alter the final representations of sentences, resulting in a drop in performance.

\section{Conclusion and Future Work}

This paper presents \frameworkname, a framework to create
baseline models well-suited for transliterated data from
existing mPLMs, without any training. We
show \frameworkname-modified models not only preserve the
ability to deal with data written in their original scripts
but also demonstrate good capability in processing
transliterations of data originally written in non-Latin
scripts. Our experiments indicate the modified models consistently outperform the original mPLMs in the transliterated evaluation. In addition, we show that \frameworkname is particularly effective for transliterated data of languages written in phonetic scripts like Cyrillic and Devanagari.
The modified models therefore serve as strong baselines for transliterated data.
% and also potentially good starting points for continued pretraining or finetuning on domain-specific (transliterated) data.

\frameworkname can have multiple uses for future work in the community. First, \frameworkname can be used to create baselines for transliterated evaluation, which does not involve any training. Second, \frameworkname can be used to modify existing models and then the modified models can be used as good starting points for continued pretraining or finetuning on domain-specific (transliterated) data.

\section*{Limitations}

Though the \frameworkname-modified models can achieve much better performance than the original mPLMs on transliterated evaluation datasets, there is still a gap between it and the performance on non-transliterated (in the original script) evaluation. We propose several explanations for this phenomenon, such as subword transliteration ambiguity and tokenization differences. These issues should be able to be alleviated by further fine-tuning or continued pretraining on transliterated data. However, this is beyond the scope of the paper, as our motivation is to create strong baselines through a simple and effective framework for modifying existing mPLMs. We would therefore expect much stronger models can be obtained by using our \frameworkname-modified mPLMs as the starting points of further training/fine-tuning, which we would leave for future exploration in the community.

Another possible limitation is that we only consider mPLMs that leverage SentencePiece Unigram tokenizers. This is due to the fact that the most recent strong mPLMs favor such choices. However, it should be very easy to extend \frameworkname to mPLMs that use other types of tokenizers. For example, mBERT \citep{devlin-etal-2019-bert} uses WordPiece subword tokenizer where the vocabulary is learned through BPE. The vocabulary keeps frequencies of subwords instead of log probabilities. Therefore, we can simply use the frequencies to replace the scores being manipulated in the Merge step of \frameworkname. We would leave this exploration in the community if other tokenizers are used in their studies for instance.

Lastly, we only try \texttt{Uroman}, a universal transliteration tool that can convert any script to Latin script. However, as we show in our analysis, the process is not optimal since it is not adapted to every single language. For example, for Chinese, the tones are removed which introduces substantial ambiguity, negatively influencing the performance. This can be improved by using better transliteration tools that are more language-specific when one does not want to cover as many languages as we do but focus on a small group of related languages.

\section*{Acknowledgements}

This research was supported by DFG (grant SCHU 2246/14-1)
and The European Research Council (NonSequeToR, grant \#740516).

\bibliography{anthology,custom}

\appendix

\section{Settings and Hyperparameters}
\seclabel{hyperparam}

\begin{table}[h]
  \small
	\centering
	\def\tablesep{0.2cm}
\begin{tabular}{
  @{\hspace{\tablesep}}l@{\hspace{\tablesep}}|
  @{\hspace{\tablesep}}r@{\hspace{\tablesep}}
  @{\hspace{\tablesep}}r@{\hspace{\tablesep}}
  @{\hspace{\tablesep}}r@{\hspace{\tablesep}}
  @{\hspace{\tablesep}}c@{\hspace{\tablesep}}
}
\toprule
 & |head| & |tail| & \#class & measure (\%) \\
  \midrule
SR-B    & 94 & 275 & - & top-10 Acc. \\
SR-T    & 70 & 28  & - & top-10 Acc. \\
Taxi1500    & 89 & 262 & 6 & F1 score \\
SIB200 & 78 & 94 & 7 & F1 score \\
NER & 89 & 75 & 7 & F1 score \\
POS & 63 & 28 & 18 & F1 score \\
\bottomrule
  \end{tabular}
  \caption{Information of the evaluation datasets and used measures. |head| (resp. |tail|): number of head (resp. tail) language-scripts according to \citet{imanigooghari-etal-2023-glot500} (a language-script is a head language if it is covered by XLM-R, otherwise it is tail) \#class: the number of the categories if it belongs to a text classification or sequence labeling task.}
  \label{tab:evaluation_info_task}
\end{table}

The basic information of each downstream task dataset is shown in
Table \ref{tab:evaluation_info_task}. The number of languages in each major script group for each dataset is shown in Table \ref{tab:evaluation_info_script}. We use the same fine-tuning hyperparameters for both transliterated evaluation (train / valid /test sets are transliterated to Latin script using \texttt{Uroman} for all languages) and non-transliterated evaluation.
We introduce the detailed hyperparameters settings in the following.

\begin{table}[t]
\setlength{\belowcaptionskip}{-0.5cm}
  \small
	\centering
	\def\tablesep{0.1cm}
\begin{tabular}{
  @{\hspace{\tablesep}}l@{\hspace{\tablesep}}|
  @{\hspace{\tablesep}}r@{\hspace{\tablesep}}
  @{\hspace{\tablesep}}r@{\hspace{\tablesep}}
  @{\hspace{\tablesep}}r@{\hspace{\tablesep}}
  @{\hspace{\tablesep}}r@{\hspace{\tablesep}}
  @{\hspace{\tablesep}}r@{\hspace{\tablesep}}
  @{\hspace{\tablesep}}r@{\hspace{\tablesep}}
  @{\hspace{\tablesep}}r@{\hspace{\tablesep}}
}
\toprule
 & Latn & Cyrl & Hani & Arab & Deva & Other & All \\
 \midrule
SR-B &290 &28 &4 &11 &8 &28 &369 \\
SR-T &64 &10 &3 &5 &2 &14 &98 \\
Taxi1500 &281 &25 &4 &8 &7 &26 &351 \\
SIB200 &117 &11 &0 &13 &6 &28 &172 \\
NER &104 &17 &4 &10 &5 &24 &164 \\
POS &57 &8 &3 &5 &3 &15 &91 \\
\bottomrule
  \end{tabular}
  \caption{The number of languages in each script group in the evaluation datasets.}
  \label{tab:evaluation_info_script}
\end{table}

\paragraph{Sentence Retrieval} For both \textbf{SR-B} and \textbf{SR-T}, we use English-aligned sentences (up to 500 and 1000 for SR-B and SR-T respectively) from languages that the Glot500 and \textsc{Furina} supports (head + tail languages). This evaluation type does not involve any parameter updates: we directly use each model to generate the sentence-level representation by averaging the contextual token embeddings at the \textbf{8th} layer \citep{jalili-sabet-etal-2020-simalign, imanigooghari-etal-2023-glot500} and then perform retrieval by sorting the pairwise cosine similarities.

\paragraph{Text Classification} For both \textbf{Taxi1500} and \textbf{SIB200}, we fine-tune sequence-level classification models with a 6-classes classification head on the English train set and then select the best checkpoint using the English validation set. We train all models using Adam optimizer \citep{ba2015adam} for a maximum of 40 epochs, with a learning rate of 1e-5 and an effective batch size of 16 (batch size of 8, gradient accumulation of 2). We use a single GTX 1080 Ti GPU for training. The evaluation is done in zero-shot transfer: we directly apply the best checkpoint to the test sets of all other languages.

\paragraph{Sequence Labeling} For \textbf{NER} and \textbf{POS}, we fine-tune token-level classification models with a suitable classification head (7 for NER and 18 for POS) on the English train set and select the best checkpoint using the English validation set. We train all models using Adam optimizer for a maximum of 10 epochs. The learning rate is set to 2e-5 and the effective batch size is set to 32 (batch size of 8, gradient accumulation of 4). The training is done on a single GTX 1080 Ti GPU. The evaluation is done in zero-shot transfer: we directly apply the best checkpoint to the test sets of all other languages.

\begin{table}
\setlength{\belowcaptionskip}{-0.4cm}
\scriptsize
\centering
\setlength{\tabcolsep}{0.45mm}{}
\begin{tabular}{lrrrrrr}
\toprule
& Latn & Cyrl & Hani & Arab & Deva & Other \\
\midrule
\multirow{2}{*}{XLM-R} & 61.0 & 54.7 & 44.1 & 63.0 & 51.8 & 51.2 \\
& 56.9 & 49.4 & 40.7 & 57.3 & 62.1 & 64.8 \\
\midrule
\multirow{2}{*}{XLM-R (Min-Merge)} & 58.5 & 54.7 & 44.1 & 63.0 & 51.8 & 51.2 \\
& 53.7 & 42.9 & 34.8 & 41.7 & 45.3 & 48.0 \\
\midrule
\multirow{2}{*}{XLM-R (Average-Merge)} & 58.3 & 54.7 & 44.1 & 63.0 & 51.8 & 51.2 \\
& 53.5 & 42.7 & 34.0 & 41.4 & 45.0 & 47.6 \\
\midrule
\multirow{2}{*}{XLM-R (Max-Merge)} & 58.2 & 54.7 & 44.1 & 63.0 & 51.8 & 51.2 \\
& 53.2 & 42.6 & 34.1 & 41.1 & 44.5 & 47.2 \\
\midrule
\midrule
\multirow{2}{*}{Glot500} & 45.7 & 39.4 & 42.7 & 42.3 & 44.1 & 44.5 \\
& 45.5 & 43.8 & 34.9 & 50.8 & 53.7 & 57.0 \\
\midrule
\multirow{2}{*}{Glot500 (Min-Merge)} & 44.9 & 39.4 & 42.7 & 42.3 & 44.1 & 44.5 \\
& 43.5 & 36.1 & 30.8 & 36.7 & 38.8 & 39.1 \\
\midrule
\multirow{2}{*}{Glot500 (Average-Merge)} & 44.9 & 39.4 & 42.7 & 42.3 & 44.1 & 44.5 \\
& 43.4 & 36.1 & 30.6 & 36.5 & 38.7 & 39.0 \\
\midrule
\multirow{2}{*}{Glot500 (Max-Merge)} & 44.8 & 39.4 & 42.7 & 42.3 & 44.1 & 44.5 \\
& 43.3 & 36.0 & 30.9 & 36.3 & 38.4 & 38.9 \\
\midrule
\midrule
\multirow{2}{*}{\textsc{Furina}} & 45.7 & 39.4 & 42.7 & 42.3 & 44.1 & 44.5 \\
& 45.5 & 43.8 & 34.9 & 50.8 & 53.7 & 57.0 \\
\midrule
\multirow{2}{*}{\textsc{Furina} (Min-Merge)} & 44.9 & 39.4 & 42.7 & 42.3 & 44.1 & 44.5 \\
& 43.5 & 36.1 & 30.8 & 36.7 & 38.8 & 39.1 \\
\midrule
\multirow{2}{*}{\textsc{Furina} (Average-Merge)} & 44.9 & 39.4 & 42.7 & 42.3 & 44.1 & 44.5 \\
& 43.4 & 36.1 & 30.6 & 36.5 & 38.7 & 39.0 \\
\midrule
\multirow{2}{*}{\textsc{Furina} (Max-Merge)} & 44.8 & 39.4 & 42.7 & 42.3 & 44.1 & 44.5 \\
& 43.3 & 36.0 & 30.9 & 36.3 & 38.4 & 38.9 \\
\bottomrule
% \bottomrule
\end{tabular}
\caption{Full tokenization performance on SR-B dataset averaged by script group. The results on non-transliterated (resp. transliterated) data are in the 
first (resp. second) row for each model variant.
}\label{tab:trans_vs_non_tokens_complete}
\end{table}

\section{Full Tokenization Performance}
\seclabel{full}

We further compare each model type by reporting their average sequence length on the SR-B dataset grouped by the scripts in Table \ref{tab:trans_vs_non_tokens_complete}. Glot500 and \textsc{Furina} have the same tokenizers, therefore, they possess identical tokenization behavior when the same merge mode is applied. We observe that \frameworkname-modified models have consistently shorter lengths on transliterated data than the original mPLM.

\section{Compete Crosslingual Transfer Results}
\seclabel{complete}

We report the complete results of the performance of all model variants on \textbf{transliterated evaluation datasets} for all tasks and languages in Table \ref{tab:srb_table1}, \ref{tab:srb_table2}, \ref{tab:srb_table3}, \ref{tab:srb_table4} (\textbf{SR-B}), Table \ref{tab:srt_table1} (\textbf{SR-T}), Table \ref{tab:taxi1500_table1}, \ref{tab:taxi1500_table2}, \ref{tab:taxi1500_table3}, \ref{tab:taxi1500_table4}(\textbf{Taxi1500}), \ref{tab:sib200_table1}, \ref{tab:sib200_table2} (\textbf{SIB200}), Table \ref{tab:ner_table1}, \ref{tab:ner_table2} (\textbf{NER}), and Table \ref{tab:pos_table1} (\textbf{POS}).

\begin{table*}
\centering
\setlength{\tabcolsep}{0.7mm}{}
\resizebox{\textwidth}{!}{
\begin{tabular}{l|rrrr|rrrr|rrrr}
    \toprule
    Language & XLM-R & XLM-R (Min-Merge) & XLM-R (Average-Merge) & XLM-R (Max-Merge) & Glot500 & Glot500 (Min-Merge) & Glot500 (Average-Merge) & Glot500 (Max-Merge) & \textsc{Furina} & \textsc{Furina} (Min-Merge) & \textsc{Furina} (Average-Merge) & \textsc{Furina} (Max-Merge) \\
    \midrule
    ace\_Latn & 5.2 & \textbf{6.4} & 6.2 & 6.2 & 46.2 & 45.2 & 48.0 & \textbf{48.4} & 64.4 & 69.0 & 69.6 & \textbf{70.8} \\
ach\_Latn & 5.2 & 5.0 & \textbf{5.4} & 5.2 & \textbf{41.6} & 40.2 & 40.0 & 41.0 & \textbf{51.6} & 50.6 & 50.2 & 50.0 \\
acr\_Latn & 3.2 & \textbf{3.8} & 3.6 & \textbf{3.8} & 24.6 & 26.0 & \textbf{27.2} & 25.8 & 46.8 & \textbf{52.8} & 52.0 & \textbf{52.8} \\
afr\_Latn & \textbf{76.6} & 75.2 & 75.0 & 75.2 & \textbf{69.2} & 65.6 & 65.4 & 67.2 & 82.2 & 81.6 & \textbf{82.6} & 82.0 \\
agw\_Latn & 5.4 & \textbf{6.6} & 6.4 & 6.2 & \textbf{34.4} & 33.4 & 32.4 & 33.0 & 53.0 & \textbf{53.4} & 45.4 & 47.2 \\
ahk\_Latn & 3.4 & 3.0 & 2.4 & \textbf{3.6} & 3.4 & 3.6 & 3.4 & \textbf{3.8} & 8.6 & 10.0 & 9.6 & \textbf{10.6} \\
aka\_Latn & \textbf{5.8} & 5.2 & 5.4 & 4.4 & 28.4 & \textbf{34.0} & 32.8 & 33.4 & 44.6 & \textbf{45.6} & 45.4 & 44.2 \\
\bottomrule
    \end{tabular}
}
    \caption{Top-10 accuracy of models on \textbf{transliterated} dataset of \textbf{SR-B} (Part I).}\label{tab:srb_table1}
\end{table*}

\begin{table*}
\centering
\setlength{\tabcolsep}{0.7mm}{}
\resizebox{\textwidth}{!}{
    \begin{tabular}{l|rrrr|rrrr|rrrr}
    \toprule
    Language & XLM-R & XLM-R (Min-Merge) & XLM-R (Average-Merge) & XLM-R (Max-Merge) & Glot500 & Glot500 (Min-Merge) & Glot500 (Average-Merge) & Glot500 (Max-Merge) & \textsc{Furina} & \textsc{Furina} (Min-Merge) & \textsc{Furina} (Average-Merge) & \textsc{Furina} (Max-Merge) \\
    \midrule
    gla\_Latn & 25.0 & 25.8 & \textbf{26.4} & \textbf{26.4} & 42.4 & 42.0 & \textbf{42.6} & 42.2 & 57.2 & 60.4 & \textbf{61.8} & 61.0 \\
gle\_Latn & 23.6 & 27.0 & 27.4 & \textbf{29.6} & 28.4 & 29.8 & 29.8 & \textbf{30.4} & 51.4 & 53.2 & \textbf{54.4} & 52.4 \\
glv\_Latn & 5.8 & 5.0 & \textbf{6.0} & 5.4 & \textbf{47.4} & 39.4 & 38.8 & 39.0 & \textbf{58.2} & 54.2 & 55.6 & 54.8 \\
gom\_Latn & 5.8 & 6.8 & \textbf{7.0} & 6.6 & \textbf{44.6} & 40.2 & 40.8 & 41.0 & \textbf{58.0} & 56.6 & 57.6 & 56.0 \\
gor\_Latn & 3.8 & 4.0 & 3.8 & \textbf{4.2} & 27.4 & 27.6 & \textbf{28.6} & 28.2 & 46.4 & 46.4 & 46.6 & \textbf{47.0} \\
grc\_Grek & 4.4 & \textbf{6.8} & 5.4 & 5.8 & 10.0 & \textbf{15.8} & 14.4 & 13.0 & 14.6 & 13.2 & \textbf{14.8} & \textbf{14.8} \\
guc\_Latn & \textbf{3.4} & 3.2 & 3.2 & 2.8 & 6.0 & 11.4 & \textbf{11.8} & \textbf{11.8} & 15.4 & \textbf{22.0} & 21.6 & 20.6 \\
\bottomrule
    \end{tabular}
}
    \caption{Top-10 accuracy of models on \textbf{transliterated} dataset of \textbf{SR-B} (Part II).}\label{tab:srb_table2}
\end{table*}

\begin{table*}
\centering
\setlength{\tabcolsep}{0.7mm}{}
\resizebox{\textwidth}{!}{
    \begin{tabular}{l|rrrr|rrrr|rrrr}
    \toprule
    Language & XLM-R & XLM-R (Min-Merge) & XLM-R (Average-Merge) & XLM-R (Max-Merge) & Glot500 & Glot500 (Min-Merge) & Glot500 (Average-Merge) & Glot500 (Max-Merge) & \textsc{Furina} & \textsc{Furina} (Min-Merge) & \textsc{Furina} (Average-Merge) & \textsc{Furina} (Max-Merge) \\
    \midrule
    mam\_Latn & 3.6 & \textbf{4.2} & 3.6 & 3.4 & \textbf{13.4} & 12.4 & 12.2 & \textbf{13.4} & 30.0 & 33.2 & 31.2 & \textbf{34.0} \\
mar\_Deva & 4.2 & 20.6 & 20.2 & \textbf{26.0} & 6.8 & 20.8 & 24.8 & \textbf{31.8} & 25.0 & 39.2 & 41.6 & \textbf{46.0} \\
mau\_Latn & 2.4 & \textbf{2.8} & 2.6 & \textbf{2.8} & \textbf{3.8} & \textbf{3.8} & 3.2 & 3.0 & \textbf{7.2} & \textbf{7.2} & 6.8 & 7.0 \\
mbb\_Latn & 3.6 & 3.4 & \textbf{4.2} & 3.0 & 30.6 & \textbf{36.2} & 35.0 & \textbf{36.2} & 47.6 & 50.8 & 51.2 & \textbf{51.4} \\
mck\_Latn & 5.2 & \textbf{5.6} & 5.0 & 5.0 & \textbf{57.8} & 53.4 & 53.8 & 54.4 & \textbf{65.0} & \textbf{65.0} & 63.4 & 64.6 \\
mcn\_Latn & 6.8 & \textbf{7.4} & 7.2 & 7.0 & \textbf{32.0} & 27.4 & 28.8 & 30.2 & 40.6 & 40.8 & 41.4 & \textbf{41.6} \\
mco\_Latn & 3.0 & 2.4 & 2.8 & \textbf{3.4} & \textbf{7.2} & 7.0 & \textbf{7.2} & 6.0 & 17.4 & \textbf{19.0} & 18.8 & 17.8 \\
\bottomrule
    \end{tabular}
}
    \caption{Top-10 accuracy of models on \textbf{transliterated} dataset of \textbf{SR-B} (Part III).}\label{tab:srb_table3}
\end{table*}

\begin{table*}
\centering
\setlength{\tabcolsep}{0.7mm}{}
\resizebox{\textwidth}{!}{
    \begin{tabular}{l|rrrr|rrrr|rrrr}
    \toprule
    Language & XLM-R & XLM-R (Min-Merge) & XLM-R (Average-Merge) & XLM-R (Max-Merge) & Glot500 & Glot500 (Min-Merge) & Glot500 (Average-Merge) & Glot500 (Max-Merge) & \textsc{Furina} & \textsc{Furina} (Min-Merge) & \textsc{Furina} (Average-Merge) & \textsc{Furina} (Max-Merge) \\
    \midrule
    sah\_Cyrl & 5.0 & \textbf{5.4} & \textbf{5.4} & 5.0 & 6.8 & \textbf{19.4} & 19.2 & 18.0 & 20.8 & 32.4 & \textbf{33.0} & 29.8 \\
san\_Deva & 3.6 & \textbf{7.2} & 5.4 & 6.6 & 10.4 & \textbf{11.0} & 9.6 & 10.4 & \textbf{20.0} & 14.4 & 12.2 & 15.0 \\
san\_Latn & \textbf{3.8} & \textbf{3.8} & 3.2 & \textbf{3.8} & 8.2 & \textbf{9.2} & 8.2 & 8.6 & \textbf{16.4} & 12.0 & 12.6 & 11.8 \\
sba\_Latn & \textbf{4.6} & 4.2 & 4.4 & \textbf{4.6} & 8.2 & 10.8 & \textbf{11.4} & 11.0 & 16.2 & \textbf{19.6} & 18.6 & 19.0 \\
seh\_Latn & 6.4 & 6.8 & 6.2 & \textbf{7.0} & \textbf{74.6} & 68.2 & 70.4 & 71.0 & \textbf{82.0} & 79.8 & 80.6 & 80.8 \\
sin\_Sinh & 4.4 & \textbf{13.6} & 12.4 & \textbf{13.6} & 4.6 & 9.4 & \textbf{11.6} & 10.2 & 14.6 & 19.6 & \textbf{21.4} & 20.2 \\
slk\_Latn & \textbf{57.4} & 56.8 & 57.0 & 54.8 & 41.0 & 42.4 & \textbf{43.4} & 42.6 & \textbf{72.2} & 69.2 & 70.0 & 69.0 \\
\bottomrule
    \end{tabular}
}
    \caption{Top-10 accuracy of models on \textbf{transliterated} dataset of \textbf{SR-B} (Part IV).}\label{tab:srb_table4}
\end{table*}
\begin{table*}
\centering
\setlength{\tabcolsep}{0.7mm}{}
\resizebox{\textwidth}{!}{
    \begin{tabular}{l|rrrr|rrrr|rrrr}
    \toprule
    Language & XLM-R & XLM-R (Min-Merge) & XLM-R (Average-Merge) & XLM-R (Max-Merge) & Glot500 & Glot500 (Min-Merge) & Glot500 (Average-Merge) & Glot500 (Max-Merge) & \textsc{Furina} & \textsc{Furina} (Min-Merge) & \textsc{Furina} (Average-Merge) & \textsc{Furina} (Max-Merge) \\
    \midrule
    afr\_Latn & \textbf{71.7} & 69.9 & 67.4 & 65.7 & \textbf{80.7} & 77.0 & 77.1 & 74.8 & \textbf{84.3} & 83.6 & 82.0 & 82.3 \\
amh\_Ethi & 10.7 & \textbf{22.0} & 20.2 & 21.4 & 13.1 & \textbf{23.2} & 19.0 & 22.0 & 24.4 & 26.8 & 28.0 & \textbf{29.2} \\
ara\_Arab & 3.4 & 19.2 & 18.7 & \textbf{23.4} & 4.2 & 25.3 & 27.0 & \textbf{31.0} & 8.0 & 17.6 & 18.8 & \textbf{20.1} \\
arz\_Arab & 6.5 & 13.6 & 12.8 & \textbf{17.0} & 8.4 & 25.2 & 27.9 & \textbf{31.4} & 11.5 & 15.5 & 18.9 & \textbf{19.9} \\
ast\_Latn & 52.8 & 52.0 & 55.1 & \textbf{59.1} & \textbf{79.5} & 78.0 & 78.7 & 78.7 & \textbf{85.0} & \textbf{85.0} & 84.3 & 84.3 \\
aze\_Latn & 27.6 & 48.3 & 47.0 & \textbf{50.1} & 52.0 & 66.6 & \textbf{68.7} & 67.8 & 68.0 & \textbf{73.6} & 71.9 & 72.9 \\
bel\_Cyrl & 11.4 & 34.0 & 33.6 & \textbf{36.0} & 16.4 & 42.0 & 44.1 & \textbf{44.9} & 48.7 & 57.5 & 57.5 & \textbf{58.3} \\
\bottomrule
    \end{tabular}
}
    \caption{Top-10 accuracy of models on \textbf{transliterated} dataset of \textbf{SR-T}.}\label{tab:srt_table1}
\end{table*}
\begin{table*}
\centering
\setlength{\tabcolsep}{0.7mm}{}
\resizebox{\textwidth}{!}{
    \begin{tabular}{l|rrrr|rrrr|rrrr}
    \toprule
    Language & XLM-R & XLM-R (Min-Merge) & XLM-R (Average-Merge) & XLM-R (Max-Merge) & Glot500 & Glot500 (Min-Merge) & Glot500 (Average-Merge) & Glot500 (Max-Merge) & \textsc{Furina} & \textsc{Furina} (Min-Merge) & \textsc{Furina} (Average-Merge) & \textsc{Furina} (Max-Merge) \\
    \midrule
    ace\_Latn & 13.1 & 16.0 & 12.4 & \textbf{16.3} & 62.8 & 62.1 & 64.1 & \textbf{65.4} & 64.6 & 67.1 & \textbf{69.1} & 68.1 \\
ach\_Latn & 8.2 & \textbf{11.6} & 9.3 & 9.0 & 42.0 & 42.8 & \textbf{44.6} & 38.4 & 50.5 & 49.2 & 51.4 & \textbf{56.3} \\
acr\_Latn & 8.9 & \textbf{13.2} & 11.0 & 12.5 & 56.8 & 57.9 & \textbf{59.9} & 54.7 & 58.8 & 58.7 & \textbf{60.0} & 59.6 \\
afr\_Latn & \textbf{67.0} & 63.1 & 65.3 & 65.7 & \textbf{62.5} & 59.2 & 59.6 & 59.9 & 63.6 & 62.3 & 61.5 & \textbf{64.9} \\
agw\_Latn & 11.1 & \textbf{15.9} & 14.6 & 14.1 & 60.2 & 59.7 & \textbf{61.9} & 59.2 & 54.8 & 58.0 & 59.5 & \textbf{60.1} \\
ahk\_Latn & 6.0 & 8.1 & 7.4 & \textbf{8.6} & 6.8 & \textbf{7.5} & 5.7 & 6.7 & \textbf{7.2} & 5.9 & \textbf{7.2} & 6.3 \\
aka\_Latn & 9.0 & \textbf{14.3} & 11.3 & 11.6 & 30.9 & \textbf{39.5} & 37.5 & 37.1 & 39.1 & 41.9 & \textbf{43.7} & 43.3 \\
\bottomrule
    \end{tabular}
}
    \caption{F1 scores of models on \textbf{transliterated} dataset of \textbf{Taxi1500} (Part I).}\label{tab:taxi1500_table1}
\end{table*}

\begin{table*}
\centering
\setlength{\tabcolsep}{0.7mm}{}
\resizebox{\textwidth}{!}{
    \begin{tabular}{l|rrrr|rrrr|rrrr}
    \toprule
    Language & XLM-R & XLM-R (Min-Merge) & XLM-R (Average-Merge) & XLM-R (Max-Merge) & Glot500 & Glot500 (Min-Merge) & Glot500 (Average-Merge) & Glot500 (Max-Merge) & \textsc{Furina} & \textsc{Furina} (Min-Merge) & \textsc{Furina} (Average-Merge) & \textsc{Furina} (Max-Merge) \\
    \midrule
    gle\_Latn & 24.2 & 28.8 & \textbf{34.6} & 28.8 & 32.6 & 35.5 & \textbf{38.3} & 34.4 & 40.1 & \textbf{40.6} & 38.7 & 39.7 \\
glv\_Latn & 9.0 & \textbf{11.7} & 10.0 & 10.0 & \textbf{44.7} & 40.3 & 43.8 & 40.3 & 44.2 & 43.7 & 42.8 & \textbf{44.7} \\
gom\_Latn & 7.4 & 10.2 & \textbf{11.5} & 9.7 & \textbf{39.1} & 34.1 & 38.5 & 37.6 & 37.0 & \textbf{38.7} & 36.8 & 34.9 \\
gor\_Latn & 15.0 & 14.4 & \textbf{15.3} & 13.3 & 49.6 & 48.8 & \textbf{51.7} & 49.3 & \textbf{59.5} & 54.8 & 56.5 & 59.2 \\
guc\_Latn & 5.6 & \textbf{9.7} & 9.2 & 7.4 & 28.5 & 39.1 & \textbf{39.2} & 36.0 & 36.1 & 41.3 & 42.5 & \textbf{44.1} \\
gug\_Latn & 10.2 & \textbf{12.7} & 12.5 & 10.5 & 34.0 & 46.6 & \textbf{46.8} & 42.4 & 33.3 & 40.8 & \textbf{42.9} & 41.9 \\
guj\_Gujr & 11.8 & 30.8 & 26.1 & \textbf{41.1} & 16.3 & 38.6 & 42.0 & \textbf{42.8} & 41.7 & 41.8 & 43.7 & \textbf{50.0} \\
\bottomrule
    \end{tabular}
}
    \caption{F1 scores of models on \textbf{transliterated} dataset of \textbf{Taxi1500} (Part II).}\label{tab:taxi1500_table2}
\end{table*}

\begin{table*}
\centering
\setlength{\tabcolsep}{0.7mm}{}
\resizebox{\textwidth}{!}{
    \begin{tabular}{l|rrrr|rrrr|rrrr}
    \toprule
    Language & XLM-R & XLM-R (Min-Merge) & XLM-R (Average-Merge) & XLM-R (Max-Merge) & Glot500 & Glot500 (Min-Merge) & Glot500 (Average-Merge) & Glot500 (Max-Merge) & \textsc{Furina} & \textsc{Furina} (Min-Merge) & \textsc{Furina} (Average-Merge) & \textsc{Furina} (Max-Merge) \\
    \midrule
    mar\_Deva & 8.7 & 36.3 & 38.3 & \textbf{40.2} & 8.5 & 42.3 & \textbf{48.0} & 45.1 & 30.5 & 39.1 & \textbf{51.9} & 48.3 \\
mau\_Latn & 5.1 & 5.5 & \textbf{5.9} & 5.0 & 4.9 & 5.6 & 5.3 & \textbf{5.8} & 4.9 & 5.3 & 5.6 & \textbf{6.8} \\
mbb\_Latn & 9.9 & 11.4 & 9.8 & \textbf{12.7} & 54.9 & 57.9 & \textbf{59.5} & 56.9 & 54.8 & 57.6 & \textbf{58.5} & 58.1 \\
mck\_Latn & 11.8 & 12.7 & \textbf{14.2} & 9.9 & \textbf{54.0} & 51.2 & 53.2 & 50.5 & 54.7 & 56.4 & 54.2 & \textbf{57.5} \\
mcn\_Latn & 8.3 & \textbf{12.1} & 10.5 & 9.2 & 39.7 & \textbf{41.9} & 41.2 & 39.9 & 31.2 & 38.4 & 36.8 & \textbf{39.9} \\
mco\_Latn & 7.6 & \textbf{11.0} & 9.0 & 8.0 & 15.5 & 18.5 & \textbf{20.6} & 18.9 & 24.0 & 22.2 & 24.6 & \textbf{25.4} \\
mdy\_Ethi & 7.3 & \textbf{10.5} & 8.5 & 9.3 & 5.1 & 28.8 & \textbf{30.2} & \textbf{30.2} & 10.2 & 33.4 & 33.9 & \textbf{34.7} \\
\bottomrule
    \end{tabular}
}
    \caption{F1 scores of models on \textbf{transliterated} dataset of \textbf{Taxi1500} (Part III).}\label{tab:taxi1500_table3}
\end{table*}

\begin{table*}
\centering
\setlength{\tabcolsep}{0.7mm}{}
\resizebox{\textwidth}{!}{
    \begin{tabular}{l|rrrr|rrrr|rrrr}
    \toprule
    Language & XLM-R & XLM-R (Min-Merge) & XLM-R (Average-Merge) & XLM-R (Max-Merge) & Glot500 & Glot500 (Min-Merge) & Glot500 (Average-Merge) & Glot500 (Max-Merge) & \textsc{Furina} & \textsc{Furina} (Min-Merge) & \textsc{Furina} (Average-Merge) & \textsc{Furina} (Max-Merge) \\
    \midrule
    rus\_Cyrl & 30.3 & 54.8 & \textbf{61.3} & 57.0 & 43.4 & 52.9 & 53.3 & \textbf{55.3} & 54.6 & \textbf{59.7} & 58.6 & 58.0 \\
sag\_Latn & 8.6 & 11.2 & \textbf{11.9} & 11.1 & 50.5 & \textbf{52.9} & 52.2 & 50.9 & 48.7 & 48.7 & 49.0 & \textbf{49.7} \\
sah\_Cyrl & 7.2 & 10.1 & \textbf{10.7} & 10.3 & 5.1 & \textbf{36.7} & 30.7 & 34.2 & 10.1 & 40.2 & \textbf{40.7} & 39.4 \\
sba\_Latn & 12.7 & \textbf{13.2} & 12.7 & 11.5 & 12.6 & 21.6 & \textbf{22.0} & 21.9 & 20.2 & 23.0 & 24.8 & \textbf{25.4} \\
seh\_Latn & 8.9 & 12.1 & \textbf{13.0} & 11.2 & 51.8 & 53.6 & \textbf{54.6} & 53.2 & 56.9 & 57.1 & 55.9 & \textbf{60.6} \\
sin\_Sinh & 9.8 & 34.7 & \textbf{40.6} & 39.8 & 6.4 & 28.5 & 30.3 & \textbf{32.6} & 13.0 & 34.4 & 35.4 & \textbf{40.8} \\
slk\_Latn & 54.7 & 57.0 & \textbf{61.5} & 59.4 & 52.4 & 55.3 & \textbf{57.9} & 55.0 & 59.7 & 62.3 & 63.4 & \textbf{64.3} \\
\bottomrule
    \end{tabular}
}
    \caption{F1 scores of models on \textbf{transliterated} dataset of \textbf{Taxi1500} (Part IV).}\label{tab:taxi1500_table4}
\end{table*}

\begin{table*}
\centering
\setlength{\tabcolsep}{0.7mm}{}
\resizebox{\textwidth}{!}{
    \begin{tabular}{l|rrrr|rrrr|rrrr}
    \toprule
    Language & XLM-R & XLM-R (Min-Merge) & XLM-R (Average-Merge) & XLM-R (Max-Merge) & Glot500 & Glot500 (Min-Merge) & Glot500 (Average-Merge) & Glot500 (Max-Merge) & \textsc{Furina} & \textsc{Furina} (Min-Merge) & \textsc{Furina} (Average-Merge) & \textsc{Furina} (Max-Merge) \\
    \midrule
    ace\_Latn & \textbf{58.2} & 56.7 & 55.7 & 56.4 & 74.7 & \textbf{75.3} & 72.8 & 73.7 & \textbf{73.6} & 72.8 & 73.0 & 70.3 \\
acm\_Arab & 14.2 & \textbf{71.5} & 67.1 & 70.8 & 16.2 & 74.9 & 75.2 & \textbf{75.4} & 15.4 & 70.7 & \textbf{74.8} & 73.1 \\
afr\_Latn & \textbf{85.3} & 84.2 & 83.4 & 84.4 & 81.1 & 81.0 & \textbf{82.1} & 81.5 & \textbf{82.4} & \textbf{82.4} & 82.1 & 81.0 \\
ajp\_Arab & 13.5 & \textbf{70.5} & 64.2 & 67.8 & 17.4 & 72.5 & 71.3 & \textbf{74.0} & 21.6 & 69.6 & \textbf{73.5} & 71.2 \\
aka\_Latn & 35.0 & 34.7 & \textbf{35.4} & 34.0 & 55.6 & \textbf{58.6} & 57.4 & 55.3 & 56.6 & 57.8 & \textbf{58.7} & 55.8 \\
als\_Latn & 79.8 & 80.8 & 80.1 & \textbf{81.6} & \textbf{81.4} & 79.8 & 79.0 & 78.9 & \textbf{81.5} & 81.2 & 81.2 & 80.1 \\
amh\_Ethi & 15.5 & \textbf{51.3} & 43.8 & 45.2 & 19.9 & \textbf{51.4} & 47.5 & 48.3 & 23.0 & 48.3 & \textbf{53.1} & 50.2 \\
\bottomrule
    \end{tabular}
}
    \caption{F1 scores of models on \textbf{transliterated} dataset of \textbf{SIB200} (Part I).}\label{tab:sib200_table1}
\end{table*}

\begin{table*}
\centering
\setlength{\tabcolsep}{0.7mm}{}
\resizebox{\textwidth}{!}{
    \begin{tabular}{l|rrrr|rrrr|rrrr}
    \toprule
    Language & XLM-R & XLM-R (Min-Merge) & XLM-R (Average-Merge) & XLM-R (Max-Merge) & Glot500 & Glot500 (Min-Merge) & Glot500 (Average-Merge) & Glot500 (Max-Merge) & \textsc{Furina} & \textsc{Furina} (Min-Merge) & \textsc{Furina} (Average-Merge) & \textsc{Furina} (Max-Merge) \\
    \midrule
    lij\_Latn & 68.8 & \textbf{70.8} & 66.7 & 68.5 & 75.7 & 76.9 & 76.2 & \textbf{77.4} & 78.1 & 77.3 & \textbf{78.5} & 77.9 \\
lim\_Latn & 70.4 & \textbf{70.9} & 69.3 & 69.2 & \textbf{76.1} & 75.4 & 74.9 & \textbf{76.1} & \textbf{78.1} & 77.1 & 77.1 & 77.0 \\
lin\_Latn & \textbf{40.2} & 37.0 & 36.7 & 37.5 & \textbf{71.3} & 71.1 & \textbf{71.3} & 69.5 & 70.5 & 68.4 & \textbf{72.5} & 69.8 \\
lit\_Latn & 81.3 & \textbf{84.1} & 81.9 & 82.1 & 79.9 & 82.3 & \textbf{83.1} & 82.3 & 83.0 & \textbf{84.3} & 84.1 & 84.0 \\
lmo\_Latn & 69.1 & \textbf{69.7} & 64.0 & 67.2 & 78.6 & \textbf{79.8} & 79.5 & 77.8 & 79.3 & \textbf{81.0} & 80.4 & 80.7 \\
ltz\_Latn & 61.1 & \textbf{62.5} & 62.4 & 61.5 & 74.9 & 74.9 & \textbf{75.7} & 74.7 & 79.1 & 78.7 & \textbf{79.9} & 78.0 \\
lua\_Latn & 38.9 & 41.1 & 39.4 & \textbf{41.5} & 57.0 & 58.1 & \textbf{59.7} & 56.1 & 55.0 & 54.8 & 55.6 & \textbf{56.5} \\
\bottomrule
    \end{tabular}
}
    \caption{F1 scores of models on \textbf{transliterated} dataset of \textbf{SIB200} (Part II).}\label{tab:sib200_table2}
\end{table*}
\begin{table*}
\centering
\setlength{\tabcolsep}{0.7mm}{}
\resizebox{\textwidth}{!}{
    \begin{tabular}{l|rrrr|rrrr|rrrr}
    \toprule
    Language & XLM-R & XLM-R (Min-Merge) & XLM-R (Average-Merge) & XLM-R (Max-Merge) & Glot500 & Glot500 (Min-Merge) & Glot500 (Average-Merge) & Glot500 (Max-Merge) & \textsc{Furina} & \textsc{Furina} (Min-Merge) & \textsc{Furina} (Average-Merge) & \textsc{Furina} (Max-Merge) \\
    \midrule
    ace\_Latn & \textbf{33.6} & 29.5 & 29.7 & 30.0 & 43.1 & 45.7 & \textbf{46.2} & 44.4 & \textbf{44.5} & 43.8 & 40.8 & 42.9 \\
afr\_Latn & 74.9 & 75.6 & \textbf{75.9} & 74.9 & 75.0 & 76.2 & \textbf{76.7} & 76.4 & 75.6 & \textbf{76.2} & 75.3 & 76.1 \\
als\_Latn & \textbf{61.7} & 56.8 & 57.8 & 55.7 & 76.8 & \textbf{81.2} & 78.4 & 80.3 & 77.6 & 79.8 & 78.5 & \textbf{82.6} \\
amh\_Ethi & 13.9 & \textbf{34.3} & 30.5 & 29.0 & 11.1 & 38.1 & \textbf{40.0} & \textbf{40.0} & 23.0 & 45.7 & 40.3 & \textbf{46.2} \\
ara\_Arab & 7.3 & 23.4 & 27.9 & \textbf{28.6} & 8.9 & 28.5 & 29.8 & \textbf{33.6} & 11.7 & \textbf{40.8} & 31.2 & 35.3 \\
arg\_Latn & 67.9 & 67.3 & 64.0 & \textbf{70.0} & \textbf{78.4} & 76.8 & 76.3 & 75.0 & 76.3 & \textbf{79.4} & 71.8 & 74.4 \\
arz\_Arab & 11.2 & 29.9 & 26.4 & \textbf{35.3} & 8.9 & 35.6 & 31.5 & \textbf{38.1} & 13.4 & 40.2 & \textbf{42.1} & 38.2 \\
\bottomrule
    \end{tabular}
}
    \caption{F1 scores of models on \textbf{transliterated} dataset of \textbf{NER} (Part I).}\label{tab:ner_table1}
\end{table*}

\begin{table*}
\centering
\setlength{\tabcolsep}{0.7mm}{}
\resizebox{\textwidth}{!}{
    \begin{tabular}{l|rrrr|rrrr|rrrr}
    \toprule
    Language & XLM-R & XLM-R (Min-Merge) & XLM-R (Average-Merge) & XLM-R (Max-Merge) & Glot500 & Glot500 (Min-Merge) & Glot500 (Average-Merge) & Glot500 (Max-Merge) & \textsc{Furina} & \textsc{Furina} (Min-Merge) & \textsc{Furina} (Average-Merge) & \textsc{Furina} (Max-Merge) \\
    \midrule
    lij\_Latn & 37.5 & 35.7 & 34.5 & \textbf{37.8} & \textbf{47.4} & 43.2 & 42.7 & 45.7 & \textbf{45.1} & 44.0 & 42.4 & 44.9 \\
lim\_Latn & 61.5 & \textbf{65.2} & 62.2 & 57.9 & 70.5 & 68.7 & 72.6 & \textbf{73.3} & 71.4 & 70.2 & 67.8 & \textbf{71.7} \\
lin\_Latn & \textbf{39.3} & 39.0 & 34.8 & 38.0 & 48.8 & 54.2 & 53.3 & \textbf{54.7} & 49.5 & 56.7 & 56.2 & \textbf{57.9} \\
lit\_Latn & 67.4 & 68.3 & \textbf{69.2} & 69.0 & 69.5 & 70.6 & 70.4 & \textbf{71.9} & 71.8 & \textbf{74.4} & 73.5 & 73.3 \\
lmo\_Latn & \textbf{72.5} & 68.2 & 68.2 & 68.7 & 70.6 & 72.4 & 71.5 & \textbf{74.6} & 70.4 & \textbf{77.4} & 72.5 & 75.1 \\
ltz\_Latn & 49.5 & 50.2 & 51.1 & \textbf{51.7} & 66.7 & \textbf{67.6} & 67.5 & 67.5 & 68.9 & \textbf{70.0} & 68.2 & 69.8 \\
lzh\_Hani & \textbf{8.3} & 5.8 & 4.8 & 6.8 & \textbf{9.0} & 7.7 & 4.3 & 6.2 & 6.6 & 9.0 & \textbf{9.6} & 6.1 \\
\bottomrule
    \end{tabular}
}
    \caption{F1 scores of models on \textbf{transliterated} dataset of \textbf{NER} (Part II).}\label{tab:ner_table2}
\end{table*}
\begin{table*}
\centering
\setlength{\tabcolsep}{0.7mm}{}
\resizebox{\textwidth}{!}{
    \begin{tabular}{l|rrrr|rrrr|rrrr}
    \toprule
    Language & XLM-R & XLM-R (Min-Merge) & XLM-R (Average-Merge) & XLM-R (Max-Merge) & Glot500 & Glot500 (Min-Merge) & Glot500 (Average-Merge) & Glot500 (Max-Merge) & \textsc{Furina} & \textsc{Furina} (Min-Merge) & \textsc{Furina} (Average-Merge) & \textsc{Furina} (Max-Merge) \\
    \midrule
    afr\_Latn & \textbf{88.8} & 87.4 & 87.3 & 87.1 & \textbf{87.8} & 86.4 & 86.4 & 86.5 & \textbf{87.7} & 87.5 & \textbf{87.7} & 87.6 \\
ajp\_Arab & 28.2 & 43.5 & \textbf{43.7} & 43.0 & 28.5 & 45.7 & \textbf{49.9} & 49.5 & 39.6 & 49.3 & \textbf{54.0} & 52.8 \\
aln\_Latn & \textbf{51.6} & 50.3 & 50.2 & 49.9 & \textbf{49.5} & 47.9 & 47.3 & 49.0 & \textbf{55.0} & 53.5 & 52.0 & 52.7 \\
amh\_Ethi & 31.7 & 48.1 & 49.5 & \textbf{50.0} & 32.8 & 46.2 & 47.8 & \textbf{51.0} & 35.7 & 48.9 & \textbf{50.4} & 49.8 \\
ara\_Arab & 18.5 & 48.9 & 51.3 & \textbf{53.9} & 23.4 & 45.4 & 51.2 & \textbf{51.6} & 38.8 & 52.7 & \textbf{56.7} & \textbf{56.7} \\
bam\_Latn & 25.7 & \textbf{27.9} & 27.6 & 27.3 & 38.4 & 43.7 & \textbf{43.9} & 42.8 & 52.4 & \textbf{53.6} & 53.0 & 52.0 \\
bel\_Cyrl & 46.6 & 79.2 & \textbf{79.4} & \textbf{79.4} & 53.5 & 76.6 & 77.8 & \textbf{78.4} & 74.6 & 81.7 & \textbf{82.2} & 82.0 \\
\bottomrule
    \end{tabular}
}
    \caption{Top-10 accuracy of models on \textbf{transliterated} dataset of \textbf{POS}.}\label{tab:pos_table1}
\end{table*}

\end{document}